%% file: main.tex
\documentclass[10pt,twocolumn,letterpaper]{article}

\input{configs/00_packages}
\input{configs/00_definitions}

\def\cvprPaperID{*}
\def\confName{CVPR}
\def\confYear{2023}
\begin{document}

%%%%%%%%% TITLE
\title{\ourtitle}

\input{configs/00_authors}

% \maketitle

\input{figtex/FIG_01_teaser}
\input{paper/SEC_00_abstract}

\input{paper/SEC_01_introduction}

\input{paper/SEC_02_related_work}

\input{paper/SEC_03_method}

\input{paper/SEC_04_experiments}
\input{paper/SEC_05_discussion}
\input{paper/SEC_06_conclusion}

\input{paper/SEC_08_ACKOWN_DISCOL}

{\small
\bibliographystyle{configs/ieee_fullname}
\bibliography{configs/main}
}
\balance
\clearpage

%\newpage
\input{paper/SEC_07_appendix}

\end{document}

%% file: configs/00_packages.tex
\usepackage[pagenumbers]{configs/cvpr} % To force page numbers, e.g. for an arXiv version

\usepackage{color}
\usepackage{amsthm}
\usepackage{wrapfig}
\usepackage{array}
\usepackage{newlfont} %
\usepackage{textcomp}
\usepackage{multirow}
\usepackage{hhline}
\usepackage{tabularx}
\usepackage{bigstrut}
\usepackage{afterpage}
\usepackage{algorithmic}
\usepackage{microtype}
\usepackage{mathtools}
\usepackage{booktabs} %
\usepackage[ruled]{algorithm2e} %
\usepackage{dingbat}

\usepackage{comment}
\usepackage{times}
\usepackage{epsfig}
\usepackage{graphicx}
\usepackage{amsmath}
\usepackage{amssymb}
\usepackage{cuted}
\usepackage{capt-of}
\usepackage{float}
\usepackage{enumitem}
\usepackage{balance}
\usepackage{bbding} % for x symbol
\usepackage[normalem]{ulem} % sout, xout

\usepackage[dvipsnames]{xcolor}
\definecolor{mypink3}{cmyk}{0, 0.7808, 0.4429, 0.1412}
\definecolor{mygray}{gray}{0.6}

\usepackage[toc,page]{appendix}

\usepackage[numbers,sort,compress]{natbib}

\usepackage[pagebackref,breaklinks,colorlinks]{hyperref}

\usepackage{color, colortbl}
\usepackage{amssymb}% http://ctan.org/pkg/amssymb
\usepackage{pifont}% http://ctan.org/pkg/pifont

\usepackage{graphicx}

%% file: configs/00_definitions.tex
\newcommand{\Fref}[1]{Fig.~\ref{#1}}

\newcommand{\tref}[1]{Tab.~\ref{#1}}
\newcommand{\eref}[1]{Eq.~(\ref{#1})}
\newcommand{\fref}[1]{Fig.~\ref{#1}}

\newcommand{\ourtitle}{MIME: Human-Aware 3D Scene Generation}

\newcommand{\modelname}{\textcolor{black}{MIME}\xspace}
\newcommand{\modelnamelong}{Mining Interaction and Movement to infer 3D Environments\xspace}

\newcommand{\datasetname}{\textsl{3D FRONT HUMAN}}
\newcommand{\bbox}{bounding box\xspace}
\newcommand{\bboxes}{bounding boxes\xspace}

\definecolor{DeltaColor}{rgb}{0.039,0.73,0.71}
\definecolor{SigmaColor}{rgb}{0.98,0.45,0.0}
\definecolor{AlphaColor}{rgb}{0,0,0.8}
\definecolor{BetaColor}{rgb}{0.8,0,0.8}
\definecolor{GammaColor}{rgb}{0.514,0.34,0.224}
\definecolor{EpsilonColor}{rgb}{0.353,0.725,0.906}
\definecolor{PurpleColor}{rgb}{0.5,0,0.7}
\definecolor{OrangeColor}{rgb}{0.914,0.541,0.0.141}
\definecolor{GreenColor}{rgb}{0.137,0.573,0.565}
\definecolor{RedColor}{rgb}{0.949,0.275, 0.224}
\definecolor{LightCyan}{rgb}{0.88,1,1}
\definecolor{Gray}{gray}{0.85}
\newcommand{\hw}[1]{\textcolor{blue}{[HW: #1]}}

\definecolor{BetaColor}{rgb}{0.8,0,0.8}

\definecolor{LightCyan}{rgb}{0.88,1,1}
\definecolor{lightgray}{rgb}{0.9,0.9,0.9}

\renewcommand{\etal}{et al.\xspace}

\renewcommand{\eg}{e.g.\xspace}

\newcommand{\threeD}{{3D}\xspace}

\newcommand{\Supmat}{\mbox{See more details in Sup.~Mat.}\xspace}

\newcommand{\qheading}[1]{\noindent\textbf{#1}}

\definecolor{GreenColor}{rgb}{0.137,0.573,0.565}

\newcommand{\colorRef}[1]{\textcolor{black}{#1}} % <-----
\usepackage[capitalize]{cleveref}
\crefname{figure}{\colorRef{Fig.}}{\colorRef{Figs.}}
\Crefname{figure}{\colorRef{Fig.}}{\colorRef{Figs.}}
\crefname{section}{\colorRef{Sec.}}{\colorRef{Secs.}}
\Crefname{section}{\colorRef{Sec.}}{\colorRef{Secs.}}
\Crefname{table}{\colorRef{Tab.}}{\colorRef{Tabs.}}
\crefname{table}{\colorRef{Tab.}}{\colorRef{Tabs.}}

\newcommand{\projectURL}{\url{https://mime.is.tue.mpg.de}}

\renewcommand{\paragraph}[1]{\medskip\noindent\textbf{#1}}

\newcommand{\myparagraph}[1]{\vspace{0.075in}\noindent\textbf{#1}}

%% file: configs/00_authors.tex
\author{
Hongwei Yi$^{1}$ Chun-Hao P. Huang$^{2*}$ Shashank Tripathi$^{1}$ Lea Hering$^{1}$ Justus Thies$^{1}$ Michael J. Black$^{1}$\\
$^{1}$Max Planck Institute for Intelligent Systems, T\"ubingen, Germany ~~~ $^{2}$Adobe Inc.\\
{\tt\small \{firstname.lastname\}@\{tuebingen.mpg.de\} chunhaoh@adobe.com} 
}

%% file: figtex/FIG_01_teaser.tex
\newcommand{\teaserCaption}{
{\bf Estimating 3D scenes from human movement.}
Given 3D human motion, e.g.~from motion capture or body-worn sensors, we reconstruct plausible 3D scenes in which the motion could have taken place. 
Our generative model is able to produce multiple realistic scenes that take into account the locations and poses of the person, with appropriate human-scene contact.
%Given humans as input (e.g., a motion sequence of one person performing daily activities in an indoor scene from PROXD \cite{hassan2019resolving}), we can generate variant realistic scenes where the input humans can have plausible interaction with.
}

\twocolumn[{
    \renewcommand\twocolumn[1][]{#1}
    \vspace{-0.5cm}
    \maketitle
    \centering
    \begin{minipage}{1.00\textwidth}
        \centering
        \vspace{-0.5cm}    
        \includegraphics[width=1.00 \linewidth]{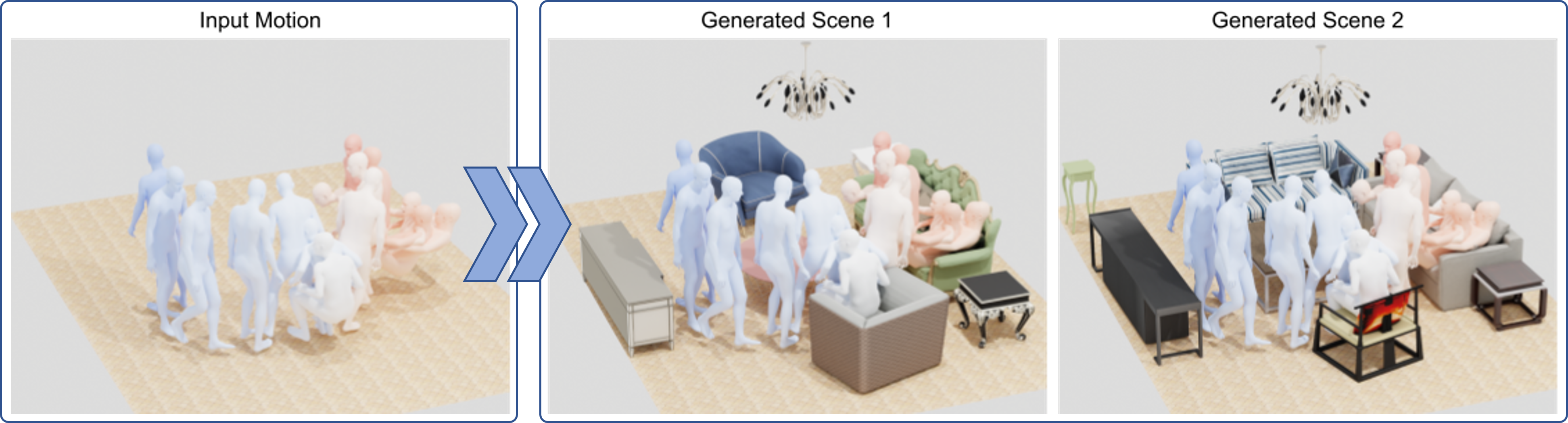}
    \end{minipage}
    \vspace{-0.1in}
    \captionof{figure}{\teaserCaption}
    \vspace{0.5cm}
    \label{fig:teaser}
}]

\def\thefootnote{*}\footnotetext{This work was performed when C.P. H. was at the MPI-IS.}

%% file: paper/SEC_00_abstract.tex
\begin{abstract}
\vspace{-0.5cm}
Generating realistic 3D worlds occupied by moving humans has many applications in games, architecture, and synthetic data creation. 
But generating such scenes is expensive and labor intensive.
Recent work generates human poses and motions given a 3D scene.
Here, we take the opposite approach and generate 3D indoor scenes given 3D human motion.
Such motions can come from archival motion capture or from IMU sensors worn on the body, effectively turning human movement into a ``scanner" of the 3D world.
Intuitively, human movement indicates the free-space in a room and human contact indicates surfaces or objects that support activities such as sitting, lying or touching.
We propose \modelname (\modelnamelong), which is a generative model of indoor scenes that produces furniture layouts that are consistent with the human movement.
\modelname uses an auto-regressive transformer architecture that takes the already generated objects in the scene as well as the human motion as input, and outputs the next plausible object.
To train \modelname, we build a dataset by populating the {\em 3D FRONT} scene dataset with 3D humans.
Our experiments show that \modelname produces more diverse and plausible 3D scenes than a recent generative scene method that does not know about human movement.
Code and data are available for research at \projectURL.
% 
% We also show that \modelname can reconstruct 3D scenes from as little information as the motion of single cell phone IMU.
% 
%Our experiments show that \modelname generalizes to different motion representations (e.g., parametric meshes or even a single root joint representation), and that \modelname outperforms state of the art in terms of generated diversity of room layouts as well as plausibility of generated objects w.r.t.~the input motion.
%
%%%%% Results/Performance:
% Through our extensive quantitative and quantitative evaluation,
%To our knowledge, our method firstly generate variant, realistic and plausible room layouts which consist of contact objects and non-contact objects, that can both support different kinds of humans.
% 

%
\end{abstract}

%% file: paper/SEC_01_introduction.tex
\section{Introduction}

Humans constantly interact with their environment.
They walk through a room, touch objects, rest on a chair, or sleep in a bed.
All these interactions contain information about the scene layout and object placement.
In fact, a mime is a performer who uses our understanding of such interactions to convey a rich, imaginary, 3D world using only their body motion.
%their body motion to convey a rich, imaginary, 3D world to the viewer. 
Can we train a computer to take human motion and, similarly, conjure the 3D scene in which it belongs?
Such a method would have many applications in synthetic data generation, architecture, games, and virtual reality.
For example, there exist large datasets of 3D human motion like AMASS \cite{AMASS:ICCV:2019} and such data rarely contains information about the 3D scene in which it was captured.
Could we take AMASS and generate plausible 3D scenes for all the motions?
If so, we could use AMASS to generate training data containing realistic human-scene interaction.
%And, what information about the 3D world can be gleaned from the motion of a single cell phone or smartwatch?
%Such devices are ubiquitous and their motion is used to recognize our activities.
%But can they also be used to reconstruct our 3D world?

To answer such questions, we train a new method called \modelname (\modelnamelong) that generates plausible indoor 3D scenes based on 3D human motion.
Why is this possible?  
The key intuitions are that (1) A human's motion through free space indicates the lack of objects, effectively {\em carving out} regions of the scene that are free of furniture.
And (2), when they are in contact with the scene, this constrains both the type and placement of 3D objects;
%that there can not exist any object; a human interacts with the 3D scene, thus a contact indicates the location of a contact object, 
e.g, a sitting human must be sitting on something, such as a chair, a sofa, a bed, etc.

To make these intuitions concrete, we develop \modelname, which 
%To this end, we present 
%\textit{MIME}: Mining Interaction and Movement of humans to infer 3D Environments.
%\textit{MIME} 
is a transformer-based auto-regressive 3D scene generation method that, given an empty floor plan and a human motion sequence, predicts the furniture that is in contact with the human.
It also predicts plausible objects that have no contact with the human but that fit with the other objects and respect the free-space constraints induced by the human motion.
%
%\cite{Paschalidou2021NEURIPS}.
%
To condition the 3D scene generation with human motion, we estimate possible contact poses using POSA~\cite{Hassan:CVPR:2021} and divide the motion in contact and non-contact snippets.
The non-contact poses define free-space in the room, which we encode as 2D floor maps, by projecting the foot vertices onto the ground plane.
%
%For contact poses, POSA predicts which vertices of the body are likely to be in contact with the scene.
%We represent these contact vertices by 3D bounding boxes.
%
The contact poses and corresponding 3D human body models are represented by 3D bounding boxes of the contact vertices predicted by POSA.
%The contact labels for these bounding boxes, such as sitting, touching, and lying, are distinguished based on the global orientation of the body poses and semantic contact body parts, see \fref{fig:data}. 
%
We use this information as input to the transformer and auto-regressively predict the objects that fulfill the contact and free-space constraints.
To train \modelname, we built a new dataset called \textit{3D-FRONT Human} that extends the large-scale synthetic scene dataset 3D-FRONT~\cite{fu20213dfront}.
Specifically, we automatically populate the 3D scenes with humans, i.e., non-contact humans (a sequence of walking motion and standing humans) as well as contact humans (sitting, touching, and lying humans). 
To this end, we leverage motion sequences from AMASS~\cite{AMASS:ICCV:2019}, as well as static contact poses from RenderPeople~\cite{Patel:CVPR:2021} scans.
%
% \textcolor{red}{how do you do this?  With POSA?  what is the process?}
%We put non-contact humans to take place in the free space on the floor plan, and we change the position of the 3D \bboxes of contact humans to interact with variant objects.

%
%At inference time, we extend ATISS \cite{Paschalidou2021NEURIPS} to generate a plausible 3D scene layout for the input motion, represented as 3D \bboxes.
At inference time, \modelname is generating a plausible 3D scene layout for the input motion, represented as 3D \bboxes.
Based on this layout, we select 3D models from the 3D-FUTURE dataset~\cite{fu20213dfuture} and refine their 3D placement based on geometric constraints between the human poses and the scene.
In comparison to pure 3D scene generation baselines like ATISS~\cite{paschalidou2021atiss}, our method generates a 3D scene that supports human contact and motion while putting plausible objects in free space.
In contrast to Pose2Room~\cite{nie2021pose2room} which is a recent pose-conditioned generative model, our method enables the generation of objects that are not in contact with the human, thus, predicting the entire scene instead of isolated objects. 
%
% And we demonstrate that our method can process different representations of human motion as input such as 3D SMPL-X bodies or even a single root joint. 
We demonstrate that our method can directly be applied to real captured motion sequences such as PROXD \cite{hassan2019resolving} \textsl{without finetuning}. 
% different representations of human motion as input such as 3D SMPL-X bodies or even a single root joint. 
% \TODO{todo}
%

\medskip
\noindent
In summary, we make the following contributions:
\begin{itemize}
    \item a novel motion-conditioned generative model for 3D room scenes that auto-repressively generates objects that are in contact with the human or avoid free-space defined by the motion.
    \item a new 3D scene dataset with interacting humans and free space humans which is constructed by populating 3D FRONT with static contact/standing poses from RenderPeople and motion data of AMASS.
\end{itemize}

\input{figtex/FIG_02_body_representation}

%% file: figtex/FIG_02_body_representation.tex
\begin{figure}
    \centerline{
    \includegraphics[width=\columnwidth, height=8cm]{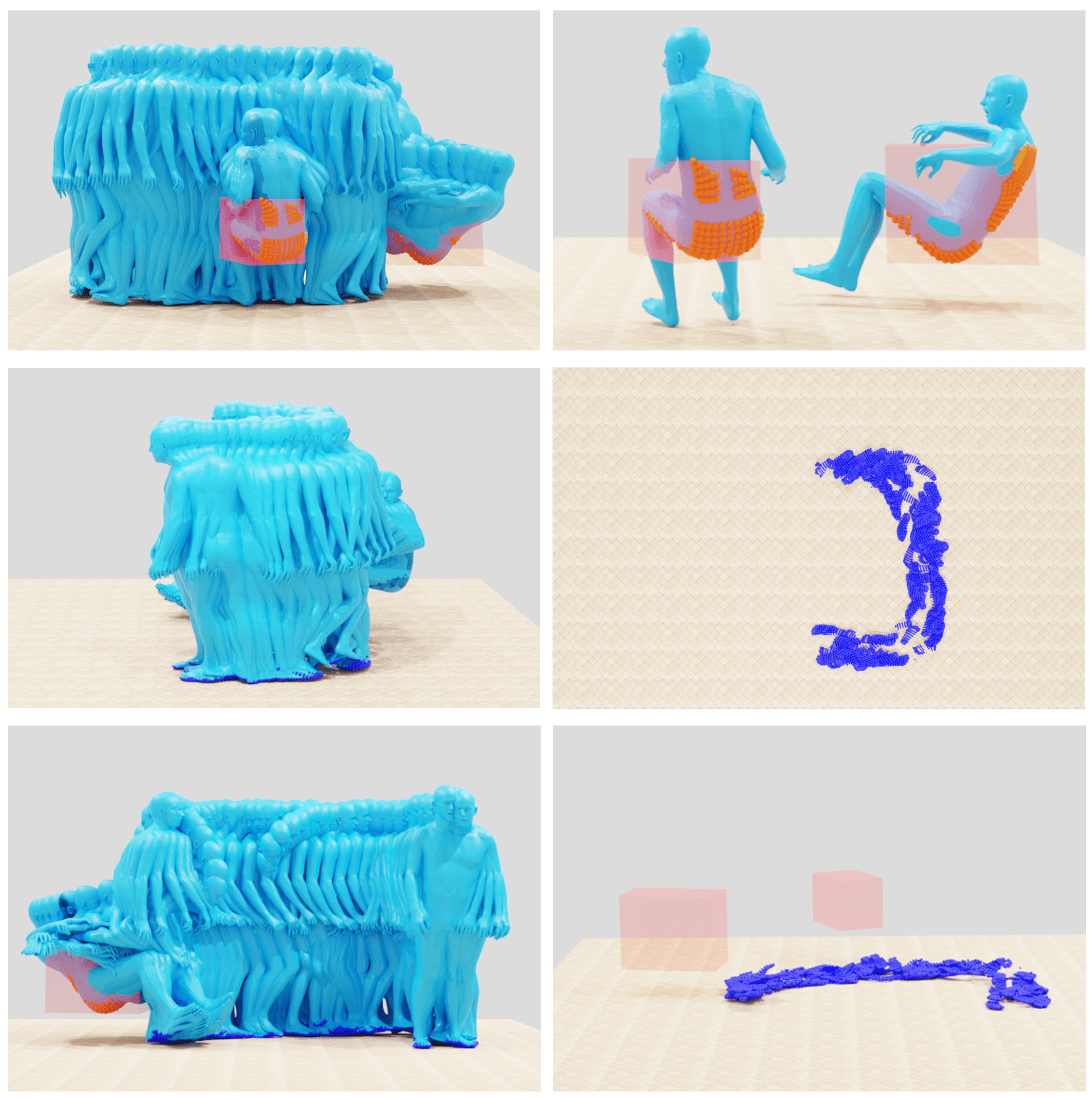}}
    %\vspace{-0.1in}
    \caption{We divide input humans into two parts: contact humans and free-space humans. 
    We extract the 3D \bboxes for each contact human, and use non-maximum suppression on the 3D IoU to aggregate multiple humans in the same 3D space into a single contact 3D \bbox (orange boxes).
    We project the foot vertices of free-space humans on the floor plane, to get the 2D free-space mask (dark blue). }
    \label{fig:data}
\end{figure}

%% file: paper/SEC_02_related_work.tex
%%%%%%%%%%%%%%%%%%%%%%%%%%%%%%%%%%%%%%%%%%%%
\input{figtex/FIG_03_architecture}

%%%%%%%%%%%%%%%%%%%%%%%%%%%%%%%%%%%%%%%%%%%%

\section{Related Work}
% \hw{Why the current datasets are not enough for our cases. Add a table for all different scenes. }

%\modelname is a {\color{red} conditional} generative model of 3D scenes given human motion inputs.
%
%It is related to unconditional generative models for scene synthesis, as well as to scene generation from human pose information.
%
%To train \modelname, we construct a synthetic scene dataset that contains human-scene interaction.
%
%We review scene synthesis methods without people, scene generation from human pose, and datasets used for the problem.

%%%%%%%%%%%%%%%%%%%%%%%%%%%%%%%%%%%%%%%%%%%%%%%%
%%%%%%%%%%%%%%%%%%%%%%%%%%%%%%%%%%%%%%%%%%%%%%%%
%%%%%%%%%%%%%%%%%%%%%%%%%%%%%%%%%%%%%%%%%%%%%%%%

\noindent\textbf{Generative Scene Synthesis (No People).}
%
%There is a series of work on indoor scene synthesis which is based on:
Most prior work on indoor scene synthesis, ignores the human and is based on
(1) procedural modeling with grammars\cite{muller2006procedural, parish2001procedural, prakash2019structured, talton2011metropolis, kar2019metasim, devaranjan2020metasim2, purkait2020sg};
(2) graph neural networks \cite{li2019grains,wang2019planit,zhou2019scenegraphnet,zhou2019scenegraphnet,luo2020end,purkait2020sg,zhang2020deep,zhang2020fast,keshavarzi2020scenegen,di2020structural};
(3) autoregressive neural networks \cite{ritchie2019fast,wang2018deep};
or (4) transformers \cite{wang2020sceneformer,Paschalidou2021NEURIPS,para2022cofs}.
Some works leverage lexical text~\cite{chang2015text} or a sentence~\cite{chang2017sceneseer} as input to guide the 3D scene synthesis. 
Fisher et al.~\cite{fisher2012example} take 3D scans as input and synthesize the corresponding 3D object arrangements.
This is extended \cite{fisher2015actsynth} to also include functionality aspects in the reconstruction.
Recently, ATISS \cite{Paschalidou2021NEURIPS} 
performs scene synthesis using a transformer-based architecture.
ATISS takes a floorplan as input and auto-regressively generates a 3D scene that is represented as an unordered set of objects.
All methods mentioned above do not take human motion into consideration to guide the 3D scene synthesis.
In contrast, we generate 3D scenes that are compatible with the humans defined by a given input motion.
Specifically, the objects in the generated scene should support the human motion (e.g., a chair or couch for sitting) and should not collide with the path of a walking human,
To this end, we build upon the auto-regressive scene synthesis architecture of ATISS~\cite{Paschalidou2021NEURIPS} and incorporate contact and free-space information into the pipeline.
% 

%%%%%%%%%%%%%%%%%%%%%%%%%%%%%%%%%%%%%%%%%%%%%%%%
%%%%%%%%%%%%%%%%%%%%%%%%%%%%%%%%%%%%%%%%%%%%%%%%
%%%%%%%%%%%%%%%%%%%%%%%%%%%%%%%%%%%%%%%%%%%%%%%%

\myparagraph{Human-aware Scene Reconstruction.}
Qi~\etal~\cite{qi2018human} propose a method that synthesizes a 3D scene based on a human's affordance map together with a spatial And-Or graph.
PiGraphs~\cite{savva2016pigraphs} learns 
a probability distribution over human pose and object geometry from interactions.
It does not model the lack of interaction, i.e.~the free space carved out by movement.
%human-object interaction snapshots from RGB-D observations, however, it does not contain any walking-human or free-space information of the room.
%
Similarly, recent methods~\cite{mura2021walk2map,nie2021pose2room} explore how to estimate a 3D scene from human behaviors and interactions.
Mura~\etal~\cite{mura2021walk2map}  predict the ``3D floor plan"  from a 2D human walking trajectory in a deterministic way. 
The approach only indicates the room layout and furniture footprints and does not model objects or contact.
Nie~\etal~\cite{nie2021pose2room} propose Pose2Room, which predicts 3D objects inside a room from 3D human pose trajectories in a probabilistic way, by learning 3D object arrangement distribution.
It only predicts contacted objects and can not generate objects in free space.
In addition, it cannot take floor plans as input. 
We find these crucial in our experiments since object arrangements are highly related to the floor plan; e.g.~some furniture is designed to go against a wall.
%

%Besides, our method does not incorporate temporary information from motion sequence, thus we can take separated multiple persons or multiple sequence of human motions as input, while Nie~\etal~\cite{nie2021pose2room} takes motion as input and sensitive to different kinds of motion sequences source.

%%%%%%%%%%%%%%%%%%%%%%%%%%%%%%%%%%%%%%%%%%%%%%%%
%%%%%%%%%%%%%%%%%%%%%%%%%%%%%%%%%%%%%%%%%%%%%%%%
%%%%%%%%%%%%%%%%%%%%%%%%%%%%%%%%%%%%%%%%%%%%%%%%

\myparagraph{Human-Scene Interaction Datasets.}
Many datasets exist for understanding humans or scenes in separation, but relatively few address humans and scenes together.
Human bodies are commonly captured using
optical markers \cite{sigal2010humaneva,h36m_pami,cmu_mocap}, 
IMU sensors \cite{vonmarcard_eccv_2018_3dpw,huang2018deep},
and multiple RGB cameras \cite{Joo_2017_TPAMI,Yu_2020_CVPR,mehta20183dv}.
See \cite{tian2022hmrsurvey} for a comprehensive review. 
These datasets contain only humans, forgoing the 3D environments which the subjects interact with,~\eg, floor plane, walls, furniture.
%
% Isolated 3D Scene Capturing:.
In contrast, real 3D scene datasets such as Matterport3D~\cite{Matterport3D}, ScanNet~\cite{dai2017scannet} and Replica~\cite{replica19arxiv} are captured primarily through time-of-flight sensors, where 
humans are excluded since only static content is reconstructed.
Consequently, despite having a large variety of scenes, they are not suitable for modeling human-scene interaction.
%

%Capturing and understanding Human-Scene Interaction.
To train \modelname, we need diverse scene arrangement given a set of sparse or continuously-moving bodies.
While recent real datasets \cite{Huang_CVPR2022,guzov2021human,hassan2019resolving,iMapper2018,bhatnagar22behave,wang2019geometric} capture both humans and environments, they fail to provide sufficient variety because the \emph{a priori} scanned scenes are static and only the subject moves.
This limits the variety of scenes that can be practically captured.
Hassan et al.~\cite{,hassan_samp_2021} use mocap to capture a person interacting with objects like chairs, sofas and tables.
They then augment the dataset by changing the size and shape of the objects and updating the human pose using inverse kinematics. The approach does not capture full scenes.
For \modelname we need a dataset with more variety.
%
%Since scanning a scene multiple time for every possible object arrangements is arguably prohibitive, real data capture is not a scalable way to build a dataset for \modelname. 
% meet our requirement
% 1) capture: RICH, BEHAVE, ~\cite{};
% But capturing dataset is time-consuming and hard to scale up.
%
Composite or synthetic datasets such as \cite{Patel:CVPR:2021, bazavan2021hspace, cai2021playing} are also widely used for human mesh recovery, but the meaningful human-scene interaction in them is fairly limited.
To our knowledge, Pose2Room \cite{nie2021pose2room} and GTA-IM \cite{caoHMP2020} are the closest to our needs.
However, they represent humans with 3D skeletons, which cannot represent realistic contact between the body surface and the scene.
Also the 
%which lack geometrically detailed human-scene interaction, and the variation of 
scene arrangement is still not rich enough to train a generative model.
Thus, we introduce a new dataset called 3D FRONT Human, which is generated by populating 3D scenes from 3D FRONT~\cite{fu20213dfront} with humans that move and interact with the scene.

%% file: figtex/FIG_03_architecture.tex
\begin{figure*}[t]
    % \ceneterline{
    % %\includegraphics[width=\textwidth]{fig/architecture.png}
    % \includegraphics[width=\textwidth]{fig/mime_pipeline.png}}
    \centerline{
    \includegraphics[width=\textwidth]{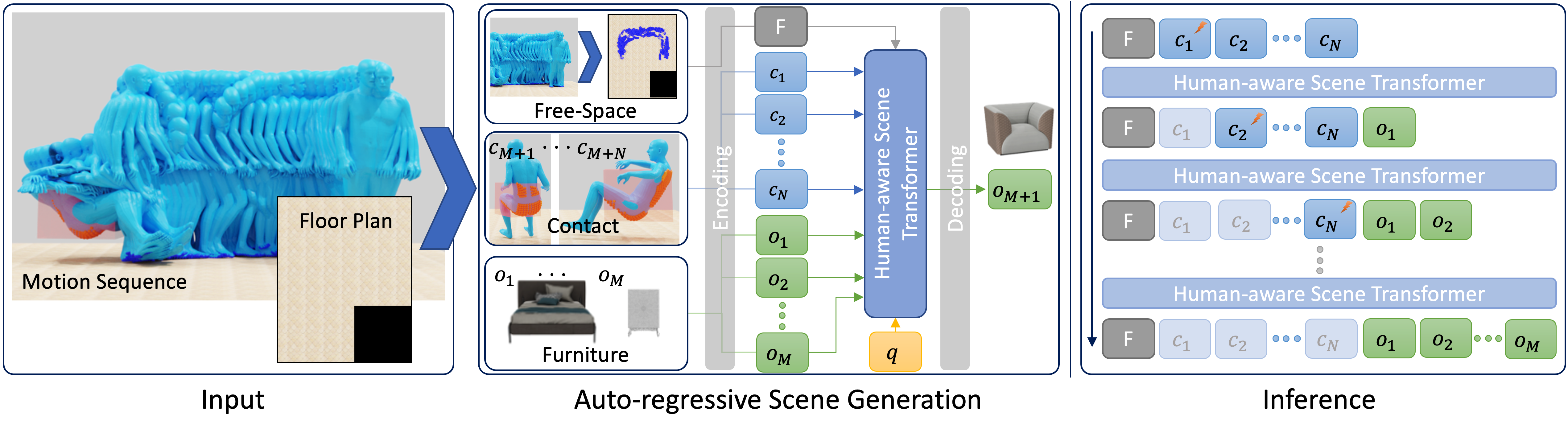}}
    
    % \vspace{-0.05in}
    \caption{Method overview. 
    % Given humans and the floor plan as input, we concatenate the free space mask $\mathcal{FS}$ with the floor plan ${\mathcal{F}}$ to extract the feature $F$ through the \textbf{layout encoder}. The contact humans $\mathcal{C}_{j=M+1}^{M+N}$, combined with the existing objects ${\mathcal{O}_{j=1}^{M}}$, are mapped into the context embedding ${T}_{j=1}^{M+N}$ by the \textbf{structure encoder}. 
    In training, our method generates object $M+1$ through a transformer encoder and a decoding module, conditioned on the free space concatenated with the floor plan, contact humans $c_{j=1}^{N}$, other existing objects ${o_{j=1}^{M}}$ and a learnable query $q$. 
    We minimize the negative log-likelihood between the distribution of the generated object $M+1$ and the ground truth.
    % The contact human which has interaction with the generated object ${O_{M+1}}$ is denoted by the indicator function $I_{j}=1, others are denoted as zeros.$
    % All extracted features along with a learnable query vector ${q}$ are passed into the \textbf{transformer encoder} to predict ${\hat{q}}$. 
    % , extracted from the \textbf{attribute extractor}, and the ground truth. 
    In inference, we start from the floor plane, the free space and input contact humans $c_{i=1}^{N}$ and assign the contact label of the first human as $1$ by default, to autoregressively generate objects. 
    At each step, we remove the contact humans that are overlapped with the previously generated object and generate next objects until the \textsl{end symbol} is generated.
    % including occupied objects and non-occupied objects, by sampling from the predicted distribution, until the end symbol. 
    % If the generated object is occupied by one of the input human, we mask the corresponding person out.
    % \JT{make sure we have the same symbols as in the text/formulas}
    }
    \label{fig:architecture}
\end{figure*}

%% file: paper/SEC_03_method.tex
\section{Method}
Given input motion of a human and an empty or partially occupied room of a specific kind (e.g., bedroom, living room, etc.) with its floor plan, we learn a generative model that can populate the room with objects that do not collide with the input humans and also support them.
To this end, we propose a human-aware autoregressive model that represents scenes as \textsl{one} unordered set of objects.
We divide the objects into two kinds, i.e.,  contact objects and  non-contact objects, based on the human-object interaction.
Contact objects are ones that humans interact with.
Non-contact objects can be placed anywhere in the free space of a room that makes semantic sense.
%in the available space in a room without any collision with the input humans, 
These objects enrich the content and potential functionality of a room.

In the following, we describe our human-aware scene synthesis model, \modelname, which consists of two components:
(1) a generative scene synthesis method based on 3D bounding boxes with object labels, and
(2) a 3D refinement method that takes 3D human-scene interactions into account to optimize the rotation and placement of the generated objects.
In \Cref{sec:data}, we detail the dataset generation process to train our model.

%in \sref{sec:arch}; Then, we illustrate more details about training and inference in \sref{sec:training}.
%Finally, we present to consider human-scene interaction losses to improve the generated 3D scene further, which has better human-scene interaction in \sref{sec:post}.

%%%%%%%%%%%%%%%%%%%%%%%%%%%%%%%%%%%%%%%%%%%%%%%%%%%%%%%%%%%%%%%%%%%%%%%%%%%%%%%%%%%%
%%%%%%%%%%%%%%%%%%%%%%%%%%%%%%%%%%%%%%%%%%%%%%%%%%%%%%%%%%%%%%%%%%%%%%%%%%%%%%%%%%%%
%%%%%%%%%%%%%%%%%%%%%%%%%%%%%%%%%%%%%%%%%%%%%%%%%%%%%%%%%%%%%%%%%%%%%%%%%%%%%%%%%%%%

\subsection{Generative Human-aware Scene Synthesis} \label{sec:arch}
Given humans $\mathcal{H}$ and a floor plan $\mathcal{F}$, 
our goal is to generate a ``habitat" $\mathcal{X} = \{\mathcal{H}, \mathcal{F}, \mathcal{S} \}$ where the 3D scene $\mathcal{S}$ can support all human interactions and motions. 
In contrast to the pure 3D scene generation methods \cite{Paschalidou2021NEURIPS, para2022cofs}, we focus on leveraging information from human motion to guide the 3D scene generation. 
To this end, we extract two types of information from the input motion and the corresponding human bodies: 
(i) contact humans $\mathcal{C}$ and 
(ii) free-space humans.
% free space humans $\mathcal{FS}=\{f_i\}_{i=1}^L$.
%
%
We use POSA \cite{Hassan:CVPR:2021}, to take posed human meshes and automatically label which of their vertices are potentially in contact with an object.
Free-space humans are those that are only in contact with the floor plane, $\mathcal{F}$.
These define a binary mask that we call free-space mask $\mathcal{FS}$, which is constructed by the union of all projected foot contact points on $\mathcal{F}$.
This free-space mask $\mathcal{FS}$ defines the region of a room that is free from objects as a human can stand and walk there.
Given all contact humans, we compute the bounding boxes of their contact vertices and keep only the non-overlapping boxes using non-maximum suppression; we denote these as $c_i$.
The collection of contact boxes is referred to as  $\mathcal{C} = \{c_i\}_{i=1}^{N}$.
Instead of storing all contact vertices of all bodies, our features are compact and encode complementary information.
The contact humans, represented by $\mathcal{C}$, indicate where to locate an object.
% where we can not put an object. 
See \fref{fig:data} top and middle rows for an illustration.
%

% Scene representation.
We represent a 3D scene $\mathcal{S}$ as an unordered set of objects, consisting of two kinds of objects based on human-object interaction.
Objects in contact with the input human are referred to as contact objects $\mathcal{O} = \{o\}_{i=1}^{N}$, while non-contact objects $\mathcal{Q} = \{q\}_{i=1}^{M}$ are without any human interaction. 
% \JT{does the number of contact objects equal the number of contact humans? what about a bed where one can sit on both sides?}\hw{In our training dataset, we make a strong assumption that one human only contains one contact object. But during inference, we use 2D IoU to factor out those contact humans which have overlap with the generated objects.}
%
Formally, a 3D scene is the union of contact and non-contact objects: $\mathcal{S}=\mathcal{O}\cup\mathcal{Q}$.
The free-space mask $\mathcal{FS}$, the floor plan $\mathcal{F}$, the contact humans $\mathcal{C}$ as well as the already existing objects $\mathcal{S}$ are input to an auto-regressive transformer model.
Each input is encoded with a respective encoder, detailed below.

The log-likelihood of the generation of scene $\mathcal{S}$ including contact objects and non-contact objects is:
\begin{equation}\label{eq1}
\log p (\mathcal{S}) = \log p(\mathcal{O}|\mathcal{F}, \mathcal{FS}, \mathcal{C}) + \log p(\mathcal{Q}|\mathcal{F}, \mathcal{FS}, \mathcal{C}).
    % log p(\mathcal{C}, Q|\mathcal{F}, \mathcal{FS}, C) = p()
    % \log p_\theta(\mathcal{X})=\sum_{i=1}^N \log \left(\sum_{\hat{\mathcal{O}} \in \pi\left(\mathcal{O}_i\right)} \prod_{j \in \hat{\mathcal{O}}} p_\theta\left(o_j^i \mid o_{<j}^i, \mathbf{F}^i\right)\right)
\end{equation}
%
% \TODO{
To calculate the likelihood of all generated contact objects $\mathcal{Q}$, 
we accumulate the likelihood of every contact object:% autoregressively in any order:
\begin{equation*}
    p(\mathcal{O}|\mathcal{F}, \mathcal{FS}, \mathcal{C}) = \sum_{\hat{\mathcal{O}} \in \pi(\mathcal{O})} 
    \prod_{j \in \hat{\mathcal{O}}} p \left(o_j \mid o_{<j}, \mathcal{F} , \mathcal{FS}, c_{\geq j} \right) ,
    % p_\theta\left(\mathcal{O}_i \mid \mathbf{F}^i\right)=\sum_{\hat{\mathcal{O}} \in \pi\left(\mathcal{O}_i\right)} \prod_{j \in \hat{\mathcal{O}}} p_\theta\left(o_j^i \mid o_{<j}^i, \mathbf{F}^i\right)
\end{equation*}
where $p \left(o_j \mid o_{<j}, \mathcal{F}, \mathcal{FS}, c_{\geq j} \right)$ is the probability of generating the $j_\text{th} $ object conditioned on the input floor plan, free-space humans, the rest of contact humans and the previously generated objects, and $\pi$ is the random permutation function for those generated contact objects in the scene.
The likelihood of all non-contact objects $\mathcal{Q}$ is computed by replacing the input contact humans with the corresponding generated contact objects. 
During the training, we remove all contact humans inside the room, thus, all contact objects $\mathcal{O}$ can be treated as non-contact objects $\mathcal{Q^{\prime}}$:
\begin{equation*}
\begin{split}
    p(\mathcal{Q}|\mathcal{F}, \mathcal{FS}, \mathcal{C}) & = p(\mathcal{Q}|\mathcal{F}, \mathcal{FS}, \mathcal{O}) \\
                                              & = p(\mathcal{Q}|\mathcal{F}, \mathcal{FS}, \mathcal{Q^{\prime}}) \\ 
                                              & = \sum_{\hat{\mathcal{Q}} \in \pi(\mathcal{Q+Q^{\prime}})} 
    \prod_{j \in \hat{\mathcal{Q}}} p \left(q_j \mid q_{<j}, \mathcal{F} , \mathcal{FS} \right) .
    % p_\theta\left(\mathcal{O}_i \mid \mathbf{F}^i\right)=\sum_{\hat{\mathcal{O}} \in \pi\left(\mathcal{O}_i\right)} \prod_{j \in \hat{\mathcal{O}}} p_\theta\left(o_j^i \mid o_{<j}^i, \mathbf{F}^i\right)
\end{split}
\end{equation*}
We follow \cite{Paschalidou2021NEURIPS} to use Monte Carlo sampling to approximate all different object permutations during training, to make our model invariant to the order of generated objects.
%
% } 
% \JT{needs to be checked, some sentences are unclear to me. e.g., notation of contact object change from O to C etc.}

%%%%%%%%%%%%%%%%%%%%%%%%%%%%%%%%%%
\paragraph{Free-Space Encoder.}
The 2D free-space mask $\mathcal{FS}$ is encoded together with the 2D floor plan $\mathcal{F}$ using a ResNet-18 \cite{he2016deep}.
The encoded feature provides the information to the transformer encoder about where an object can be placed.
% 
%We follow ATISS \cite{Paschalidou2021NEURIPS} to choose ResNet-18 \cite{he2016deep} as the free-space encoder.

%%%%%%%%%%%%%%%%%%%%%%%%%%%%%%%%%%
\paragraph{Contact Encoder.}
We represent the contact humans as 3D \bboxes, which consist of the contact label $I$, the contact class category $k$ (sitting, touching, lying), the translation $t$, the rotation $r$, and the size $s$. 
During generation of a scene, we set the contact label $I$ of one contact human to $1$ while the others are labeled $0$.
This label highlights the contribution of the specific contact human to the next generated contacted object.
Note that we remove contact humans from the input set if they are already in contact with an existing object in the scene.
%
%
%The contact label $I_{j}$ of the $j=M+1$ contact human which contacts with the generated object $o_{M+1}$ has a positive label. \JT{why do the indices need to match between the contact human and the generated object?}
%The rest of the contact humans $j>M+1$ have zero contact labels.
%This label highlights the contribution of the specific contact human to the corresponding generated contacted object.
%
Otherwise, we encode the $j_{th}$ input contact human by applying:
\begin{equation*}
    E_{\theta}: (I_j, k_j, t_j, r_j, s_j) \rightarrow (I_j, \lambda(k_j), p(t_j), p(r_j), p(s_j)),
\end{equation*}
where $\lambda\left(\cdot\right)$ is a learnable embedding for the contact class category $k$, 
and $p\left(\cdot \right)$~\cite{vaswani2017attention} is the positional encoding for the translation $t$, rotation $r$ and size $s$.
%
%The contact encoder $E_{\theta}$ transforms the attribute of each contact human $\mathcal{H}_{j}$ to the context embedding $T_{j}$, where $j \in [M+1, M+N]$:

%%%%%%%%%%%%%%%%%%%%%%%%%%%%%%%%%%
\paragraph{Furniture Encoder.}
The furniture encoder computes the embedding of existing objects in the room:
\begin{equation*}
    E_{\theta}: (I_j=0, k_j, t_j, r_j, s_j) \rightarrow (0, \lambda(k_j), p(t_j), p(r_j), p(s_j)).
\end{equation*}
Note that the furniture encoder is sharing the same weight as the contact encoder. 
The contact labels of the objects are all zero, where $j \in [1, M]$.
 % and corresponding object class category $k$ labels are used.

%%%%%%%%%%%%%%%%%%%%%%%%%%%%%%%%%%
\paragraph{Scene Synthesis Transformer.}
We pass the free-space feature $F$, context embedding $T_{i=1}^{M+N}$, and a learnable query vector $q \in \mathbb{R}^{64}$ into a transformer encoder $\tau_{\theta}$ \cite{vaswani2017attention, devlin2018bert} without any positional encoding \cite{vaswani2017attention}, to predict the feature $\hat{q}$ that is used to generate the next object: 
\begin{equation*}
    \tau_{\theta}(F, T_{i=1}^{M+N}, q) \rightarrow \hat{q}.
\end{equation*}
To decode the attribute distribution $(\hat{k}, \hat{t}, \hat{r}, \hat{s})$ of the generated object ${o_{M+1}}$ from $\hat{q}$, we follow the same design from ATISS~\cite{paschalidou2021atiss}.
Specifically, we employ an MLP for each attribute in a consecutive fashion.
Given $\hat{q}$, we first predict the class category label $\hat{k}$, then we predict the $\hat{t}$, $\hat{r}$ and $\hat{s}$ in this specific order, where the previous attribute will be concatenated with the input $\hat{q}$ for the next attribution prediction.
%
%We predict the object attributes following one MLP for each attribute in an interative way. \JT{is attribute distribution and object attribute the same here? sentence is also not very clear}
%We predict the class category $\hat{k}$ at first from $\hat{q}$ with one MLP, then we iteratively predict the $\hat{t}$, $\hat{r}$ and $\hat{s}$ in this specific order, where the previous attribute will be also concatenated with the input $\hat{q}$ for next attribution prediction.

% We use the same attribute extractor in ATISS \cite{paschalidou2021atiss} to extract the attribute distribution .

 % Please refer to ATISS \cite{paschalidou2021atiss} for more details.
% } 
% \JT{revise: it is the backbone of our method, so it should be very clear what we are doing here. Do not refer to ATISS!}

% \TODO{add more specific details about this. The content should be self-included.}
% The structure encoder, structure encoder, and attribute extractor are all following our baseline, please refer to \cite{} for more details.

\input{figtex/FIG_04_post_process}

%%%%%%%%%%%%%%%%%%%%%%%%%%%%%%%%%%%%%%%%%%%%%%%%%%%%%%%%%%%%%%%%%%%%%%%%%%%%%%%%%%%%
%%%%%%%%%%%%%%%%%%%%%%%%%%%%%%%%%%%%%%%%%%%%%%%%%%%%%%%%%%%%%%%%%%%%%%%%%%%%%%%%%%%%
%%%%%%%%%%%%%%%%%%%%%%%%%%%%%%%%%%%%%%%%%%%%%%%%%%%%%%%%%%%%%%%%%%%%%%%%%%%%%%%%%%%%

\subsection{Training and Inference.} \label{sec:training}
% \TODO{
We train our model on the training set of \textsl{\datasetname}, by maximizing the log-likelihood of each generated scene $\mathcal{S}$ in \eref{eq1}.
During training, we select a human-populated scene in \textsl{\datasetname} and add a random permutation $\pi\left(\cdot \right)$ on all $\mathit{N}$ contact and $\mathit{M}$ non-contact objects. 
% \JT{is N the number of humans?}
We randomly select the $m_\textsl{th}+1$ as the generated object, where $m \in [0, N+M]$. 
Note that, $m=0$ represents an empty scene, while $m=\mathit{N+M}$ indicates the generated scene is already full and the class label of the predicted object is an extra \textsl{end symbol}.
Our model  predicts the attribute distribution of the generated object, conditioned on the floor plane $\mathcal{F}$, free space $\mathcal{FS}$, previous $m$ objects and contact humans $\mathcal{C}$; see \fref{fig:architecture}. %$\mathcal{C}_{i\geq m+1}$, see in \fref{fig:architecture}. 
% 
% We equally input or drop out all contact humans into the input, to make our model can generate contact objects for contact humans or generate non-contact objects in the available regions carved out by the free-space humans.
To enable our model to generate both contact objects and non-contact objects, we make a data augmentation for adding input contact humans or dropping them out in equal frequency. 
% we equally input contact humans or drop them out in the input. 

%
During inference, we start from an empty floor plane $F$ with input humans including free-space humans $\mathcal{FS}$, and contact humans $\mathcal{C}$. 
We autoregressively sample the attribute of the next generated object to put one object into a scene. 
By default, we set the contact label of the first contact human to $1$, and the rest are $0$.
After each generation step, we remove contact humans that are already in contact, by computing the 2D IoU of the human \bbox and the generated object by projecting them on the ground plane. Specifically, if the IoU is larger than $0.5$, we remove the contact human from the input.
%If the generated object shares the common space with the first contact human, i.e., with 2D IoU large than 0.5, we remove this human in the next step for generating the next object. 
%
%If the generated object does not interact with any humans, we just append it to the input, as shown in \fref{fig:architecture}. 
Once the $\textsl{end symbol}$ is generated, the generated scene is finished. 
% } \JT{needs to be revised, not easy to read}

%%%%%%%%%%%%%%%%%%%%%%%%%%%%%%%%%%%%%%%%%%%%%%%%%%%%%%%%%%%%%%%%%%%%%%%%%%%%%%%%%%%%
%%%%%%%%%%%%%%%%%%%%%%%%%%%%%%%%%%%%%%%%%%%%%%%%%%%%%%%%%%%%%%%%%%%%%%%%%%%%%%%%%%%%
%%%%%%%%%%%%%%%%%%%%%%%%%%%%%%%%%%%%%%%%%%%%%%%%%%%%%%%%%%%%%%%%%%%%%%%%%%%%%%%%%%%%

\subsection{3D Scene Refinement} \label{sec:post}

The generated scene from our model is represented with 3D \bboxes.
Based on the \bbox size and class category label, we retrieve the closest mesh model from 3D FUTURE~\cite{fu20213dfuture}.
To improve the human-scene interaction between the generated scenes and input humans, we apply the collision loss and the contact loss from MOVER \cite{yi2022mover} to refine the object position, as can be seen in \fref{fig:post-process}.
We calculate a unified SDF volume and accumulate all contact vertices for all humans in the 3D space, and jointly optimize the object alignment to improve human-object contact and resolve 3D interpenetrations between humans and the scene.
The MOVER contact loss weight and the collision loss weight are $1e5$ and $1e3$ respectively. 
% \JT{check weights}

% \clearpage

%%%%%%%%%%%%%%%%%%%%%%%%%%%%%%%%%%%%%%%%%%%%%%%%%%%%%%%%%%%%%%%%%%%%%%%%%%%%%%%%%%%%
%%%%%%%%%%%%%%%%%%%%%%%%%%%%%%%%%%%%%%%%%%%%%%%%%%%%%%%%%%%%%%%%%%%%%%%%%%%%%%%%%%%%
%%%%%%%%%%%%%%%%%%%%%%%%%%%%%%%%%%%%%%%%%%%%%%%%%%%%%%%%%%%%%%%%%%%%%%%%%%%%%%%%%%%%
%%%%%%%%%%%%%%%%%%%%%%%%%%%%%%%%%%%%%%%%%%%%%%%%%%%%%%%%%%%%%%%%%%%%%%%%%%%%%%%%%%%%
%%%%%%%%%%%%%%%%%%%%%%%%%%%%%%%%%%%%%%%%%%%%%%%%%%%%%%%%%%%%%%%%%%%%%%%%%%%%%%%%%%%%
%%%%%%%%%%%%%%%%%%%%%%%%%%%%%%%%%%%%%%%%%%%%%%%%%%%%%%%%%%%%%%%%%%%%%%%%%%%%%%%%%%%%

\section{Dataset Generation of 3D FRONT HUMAN}
\label{sec:data}
To enable 3D scene generation from humans, we need a dataset that consists of large numbers of rooms with a wide variety of human interactions.
%
%However, such a dataset does not exist.
%
Since no such dataset exists, we generate a new synthetic dataset by populating the 3D rooms in the 3D FRONT \cite{fu20213dfront} with interactive humans. 
% \TODO{Add data preprocess in ATISS.}
%
%, named \textsl{\datasetname}
We name the resulting dataset \textsl{\datasetname}.
To populate the rooms of 3D FRONT with people, we insert humans with contact and humans that stand or walk in free space, as shown in \fref{fig:representation}.
We represent people with the SMPL-X model~\cite{SMPL-X:2019} and add contact humans from RenderPeople \cite{Patel:CVPR:2021} by randomly assigning plausible interactions to different contactable objects in the room.
Specifically, we allow for three types of contact interactions: touching, sitting, and lying. 
In \fref{fig:representation}~(bottom), we put a lying down person on a bed, and multiple humans interact with a nightstand or wardrobe.
%Note that, getting better plausible body poses is out of our scope, we only change the position of the 3D contact \bbox, The bodies are used for visualization only.
% 
In the free space, we put a random number of static standing people and add multiple walking motion clips from AMASS \cite{AMASS:ICCV:2019} with random start positions and directions to the scene, and remove humans that intersect with objects in the scene.

% We also place a short walking motion clip multiple times in the free space , see b) 

%
% \TODO{The standing and walking humans are ......}\JT{information about the standing and walking humans is missing}

%We design two different ways to occupy the free space of each room with humans. 
%We can either put different numbers of static standing/walking humans in the free space to mimic multiple static humans inside a room at the same time, as shown in a) \fref{fig:representation}. Or we place a short walking motion clip multiple times in the free space with random start positions and directions to mimic a human walking around a room, see b) \fref{fig:representation}.

%Secondly, we put different contact humans, represented by SMPL-X model \cite{SMPL-X:2019} from RenderPeople \cite{Patel:CVPR:2021}, to have plausible interactions with different contactable objects in each room.
% 
%We use POSA \cite{Hassan:CVPR:2021} to extract the contact vertices and get the 3D \bbox for those contact vertices to represent the contact human. Then we put those contact \bboxes in plausible positions to interact with the objects.
% 
%Specifically, we divide contact humans into three contact kinds: touching, sitting, and lying. 
%In \fref{fig:representation}(c), we put a lying down person on a bed, multiple touching humans interacting with a nightstand or a wardrobe. 
% 
%
% 
%\Supmat~for putting contact humans in other kinds of rooms.

\input{figtex/FIG_02_1_data}

%% file: figtex/FIG_04_post_process.tex
\begin{figure}[t!]
    \centerline{
    \includegraphics[width=\columnwidth]{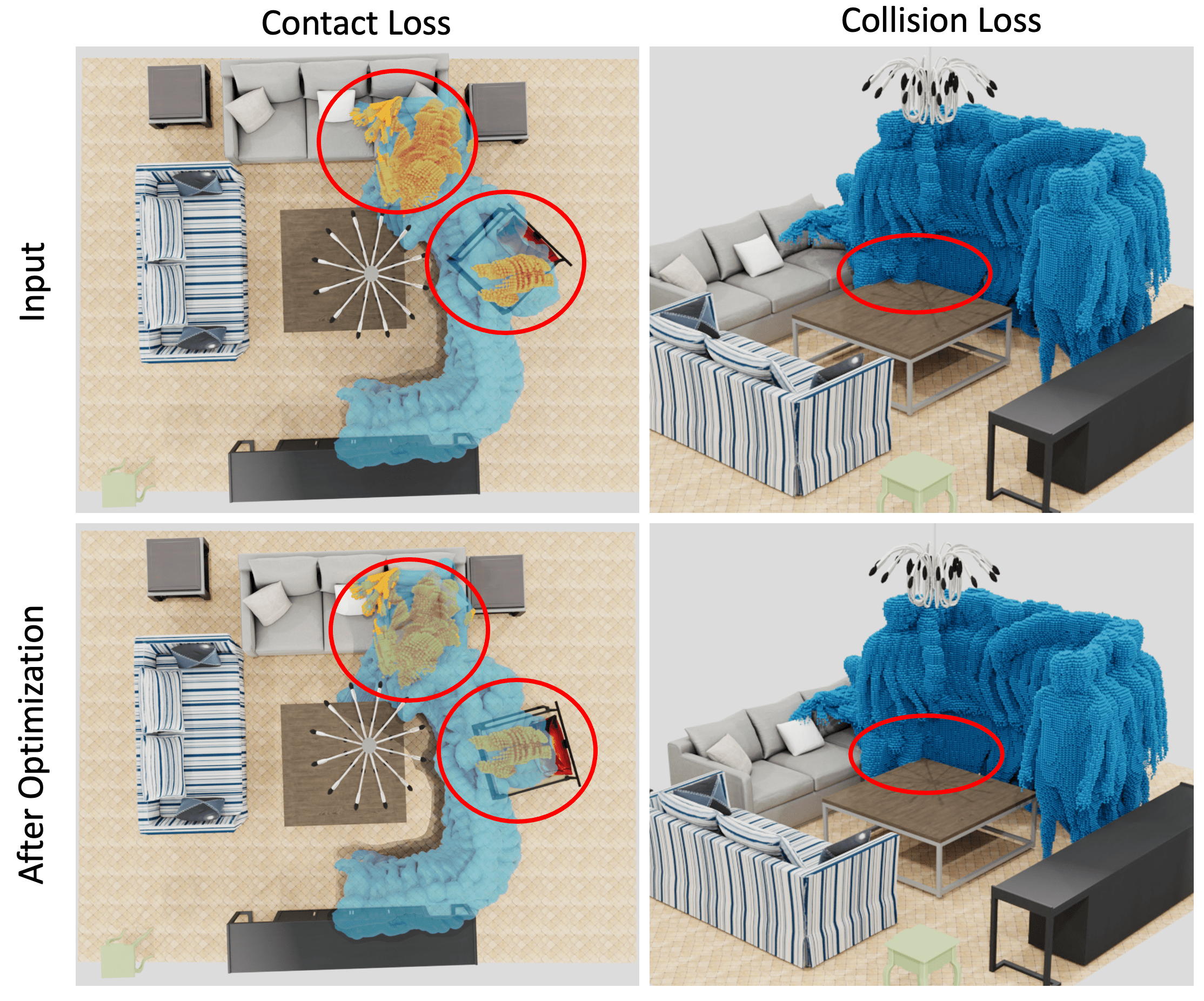}}
    %\vspace{-0.1in}
    \vspace{0.1in}
    \caption{
        Scene refinement with the collision and contact loss from MOVER \cite{yi2022mover}. In contact loss, all contact vertices (orange color) are accumulate from all bodies into 3D space and the sofa and chair are refined by minimize the one-directional Chamfer Distance with the contact vertices. In collision loss, we compute one uniform SDF volume for all bodies, where the inside of bodies are denoted as blue voxels. The table is optimized with it.
    }
    \label{fig:post-process}
\end{figure}

%% file: figtex/FIG_02_1_data.tex
\begin{figure}[t!]
    \centerline{
    \includegraphics[width=\columnwidth]{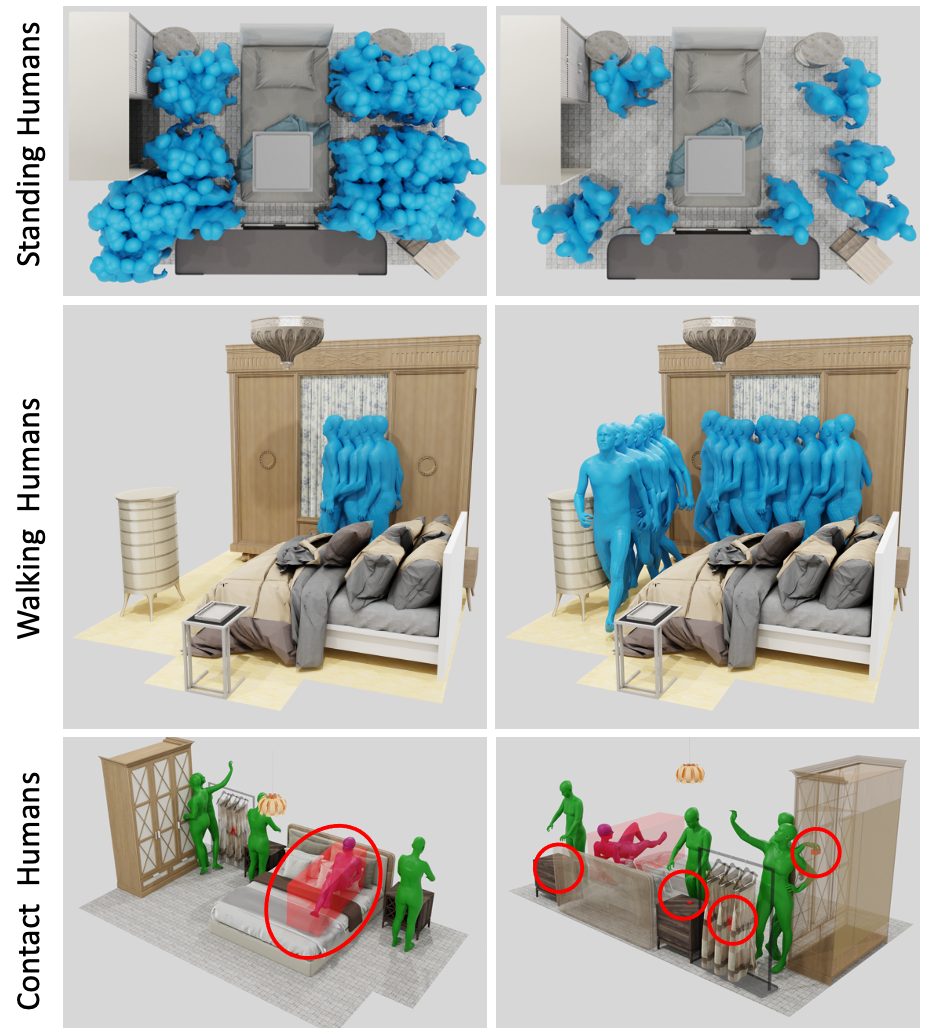}}
    %\vspace{-0.1in}
    \caption{The illustration of populated 3D scenes in \textsl{\datasetname}. 
    Given a room, we put random numbers of static ``standing" people and add multiple ``walking" motion sequences with variant start positions and directions in the free space. 
    % Alternatively, we place in the free space. 
    We also put various ``contact humans" into the scene so that their interaction with the objects makes sense, e.g., ``touching" and ``lying".}
    \label{fig:representation}
\end{figure}

%% file: paper/SEC_04_experiments.tex
\input{figtex/FIG_05_qualitative}

\input{tab/TAB_01_eval}

\input{figtex/FIG_06_00_comparison.tex}

\section{Experiments}
We qualitatively and quantitatively evaluate our method and compare with two baselines.
Specifically, we compare to the 3D scene generation method ATISS~\cite{Paschalidou2021NEURIPS} and the human-aware scene reconstruction method Pose2Room~\cite{nie2021pose2room}.
% 
%We further show several interactive cases that are enabled by our method, which our baseline totally fails.
% 
%Our method can also generate a 3D scene for the data only from a phone, which opens the possibilities to understand the environments by exploring information from wearable devices.

\paragraph{Evaluation Datasets.}
Our human-populated dataset \textsl{\datasetname} contains four room types: 1) $5689$ bedrooms, 2) $2987$ living rooms, 3) $2549$ dining rooms and 4) $679$ libraries.
We use  $21$  object categories for the bedrooms, $24$ for the living and dining rooms, and $25$ for the libraries. 
We independently train our model four times on the four kinds of rooms. 
Following our baseline ATISS \cite{Paschalidou2021NEURIPS}, for each kind of room, we split the data  \text{80\%}, \text{10\%}, \text{10\%} into training, validation and test sets.
We train and validate \modelname on the training and validation sets respectively, and evaluate it on the test set.
\Supmat
Since ATISS \cite{Paschalidou2021NEURIPS} does not provide a pretrained model, we retrain it with the official code$^\text{1}$ following the same training strategy on the original 3D FRONT dataset as one of our baseline. 
\def\thefootnote{1}\footnotetext{https://github.com/nv-tlabs/ATISS/commit/6b46c11.}

To evaluate the effectiveness and generalization of our method, we test \modelname on a real RGB-D motion captured dataset PROX-D \cite{hassan2019resolving} and compare it with Pose2Room \cite{nie2021pose2room}.
Pose2Room needs a sequence of human motions that are in contact with objects. Our \datasetname ~does not provide these interactive human-object motions, so we cannot enable fine tune and evaluate Pose2Room on \datasetname.
%
% \JT{give a reason why we are not able to have a comparison to pose2room on your other dataset!}
% another human-scene interaction synthetic dataset HUMANISE \cite{wang2022humanise} where the realistic motions are from a VICON-based captured dataset AMASS \cite{AMASS:ICCV:2019}.
 % See more details in \supmat

% \paragraph{Baseline.}
% We 

\paragraph{Evaluation Metrics.}
We compare \modelname with the baselines in two different ways: (i) the plausibility between human-scene interaction and (ii) the realism of the generated scenes only. 
We propose a \textsl{interpenetration loss} ($\downarrow$) to evaluate the collision between the generated objects and the free space, through computing the ratio of the violated free space by the 2D projection of the generated objects:
\begin{equation*}
    L_\text{inter} = \left(\sum_{j=1}^{M}\sum_{p \in O_{j}}\mathcal{FS}(p)\right) / \sum_{p \in \mathcal{FS}}\mathcal{FS}(p),
\end{equation*}
where $p$ denotes each pixel on the floor plane image.
We calculate the \textsl{2D IoU} and \textsl{3D IoU} between generated objects and input contact \bboxes to measure the human-object interaction. 
%
%Since humans are only partially in contact with an object, we replace the calculated union region with the region of the contact human.
To evaluate the realism and diversity of generated scenes, we follow common practice \cite{Paschalidou2021NEURIPS, zhang2020fast} and calculate the FID \cite{heusel2017gans} (at $256^{2}$ resolution) score between bird-eye view orthographic projections of generated scenes and real scenes from the \textsl{test} set, as well as the category KL divergence.
We compute the FID score 10 times and report the mean and variance of it.
All these evaluation experiments are conducted on the \textsl{test} split of the \datasetname~dataset.
%
% our method To evaluate the "fullfillness" of the generated scenes to input contact humans,  The number of input contact humans, 

\subsection{Human-aware Scene Synthesis.}
In \fref{fig:qualitative}, we visualize the ability of our method to generate plausible 3D scenes from input motion and floor plans for different kinds of rooms; we also show our baseline methods for comparison.
See Sup.~Mat.~for more examples.
%In \fref{fig:qualitative}, we qualitatively compare our generated scenes with our baselines. 
%
Note that the original ATISS~\cite{Paschalidou2021NEURIPS} model generates a 3D scene only based on the floor plan, without taking the humans into account.
Thus, generated scenes from ATISS violate  free space constraint and are not consistent with the human contact.
For a more fair comparison, we extend ATISS to take information about the human motion as input.  Specifically, we adapt the 2D input floor plan to also contain the free space information of the walking and standing humans.
However, ATISS with input free space still generates objects in free space, while also generating implausible object configurations such as the white closet inside the bed (\fref{fig:qualitative}, top).
In contrast, \modelname generates plausible 3D scenes that have less interpenetration with the free space and support interacting humans; e.g.~a bed beneath a lying person and a chair under a sitting person.
The observations in the qualitative comparison are also confirmed by a quantitative evaluation in \tref{tab:eval}.
\modelname achieves significant improvements on human-scene interaction evaluation metrics compared with ATISS.
Note, since our scene generation is constrained by the input motion, the diversity scores (FID, KL divergence) are lower than of ATISS, which is not human-aware.
This is not a failure/limitation of \modelname.

\input{tab/TAB_02_PROX.tex}

% \newpage
To evaluate the generalization of our method, we test it on a real dataset of human motion.
% Here we use the PROXD~\cite{hassan2019resolving} dataset \textit{without finetuning}, and use the motions to generate scenes. 
We consider the PROXD~\cite{hassan2019resolving} dataset and the 3D bounding box annotation from \cite{yi2022mover}. 
We use it \textit{without finetuning}, and use the motions to generate scenes. 
We compare our method with Pose2Room~\cite{nie2021pose2room}, which predicts 3D objects from a motion sequence of 3D skeletons.
Note that Pose2Room can only predict contact objects, it does not predict an entire scene which is the goal of our method.
\Fref{fig:cmp_on_prox} presents a qualitative comparison of the methods and we report the quantitative metrics in \tref{tab:cmp_on_prox}. 
Specifically, we compute the mean average precision with 3D IoU $0.5$ (mAP@0.5) to evaluate the 3D object detection accuracy for those contact objects only.
Both methods are probabilistic generative models to predict the object attribute distribution. Following Pose2Room, we use the same $5$ input motions and sample 10 scenes for each motion sequence, and report the mean value of it. 
% \JT{we do this only on 10 input motions or you have X input motions and predict 10 possible scenes?}
%
Our method achieves better 3D object detection accuracy compared to Pose2Room~\textsl{without pretraining}.
% After applying refinement with the collision loss and contact loss, our method achieve comparable 3D object detection performance with Pose2Room \textsl{with pretraining} on the PROX dataset. \JT{its somewhat a weird comparison, since your method is not trained on prox, right? why is the refinement so important?}

% Visualize some figures.
% Qualitative comparison. And generalization. In \fref{fig:qualitative}

% We compare our method with Pose2Room \cite{nie2021pose2room} on PROX \cite{hassan2019resolving, yi2022mover}, as shown in \fref{fig:cmp_on_prox}.

% \TODO{Use Meshlab figures as first version rendering.}
% \paragraph{Scene generation.} 

% \input{figtex/FIG_06_02_application}
% \subsection{Application.}
% See in \fref{fig:application}.

% \paragraph{Scene completion.}
% Given existing multiple objects, we put a motion sequence inside, the scene will complete.

% \paragraph{Object Rearrangement: object failure detection and correction.}

% Can we recognize where this scene can not support current input bodies?
% We can check which contact body is not occupied and which objects are violated with input free bodies.
% Then we regenerate a new scene for them.

% \paragraph{Object suggestion.}
% Given existing scenes, we put a contact human in free space, we tend to generate a contacted object with it. 

\subsection{Ablation Study on Input Humans}
In \fref{fig:diversity}, we evaluate the influence of the density of free-space humans, and the number of contact humans, that we provide as input to 
\modelname.
We observe that \modelname generates contact objects according to the number of contact humans and, as the density of free-space humans increases, \modelname generates fewer objects in scenes.  
This is as expected.
% \JT{that sounds obvious; is there any insight here?}

\input{figtex/FIG_06_01_diversity}

% 1. with and with/out Indicator Contact Flag.

% \subsection{Failure Cases.}

% \input{figtex/FIG_07_iphone}
% \paragraph{Variant Input Humans Format.}
% See in \fref{fig:iphone}
% different kinds of human motions.

% iphone;

% \subsection{Perceptual Study. [Maybe not necessary?]}

% \input{tab/TAB_02_perceptual}

% See \tref{tab:perceptual}.
% One week.

% \clearpage

%% file: figtex/FIG_05_qualitative.tex
\begin{figure*}[h]
    \centering
    \includegraphics[width=\textwidth]{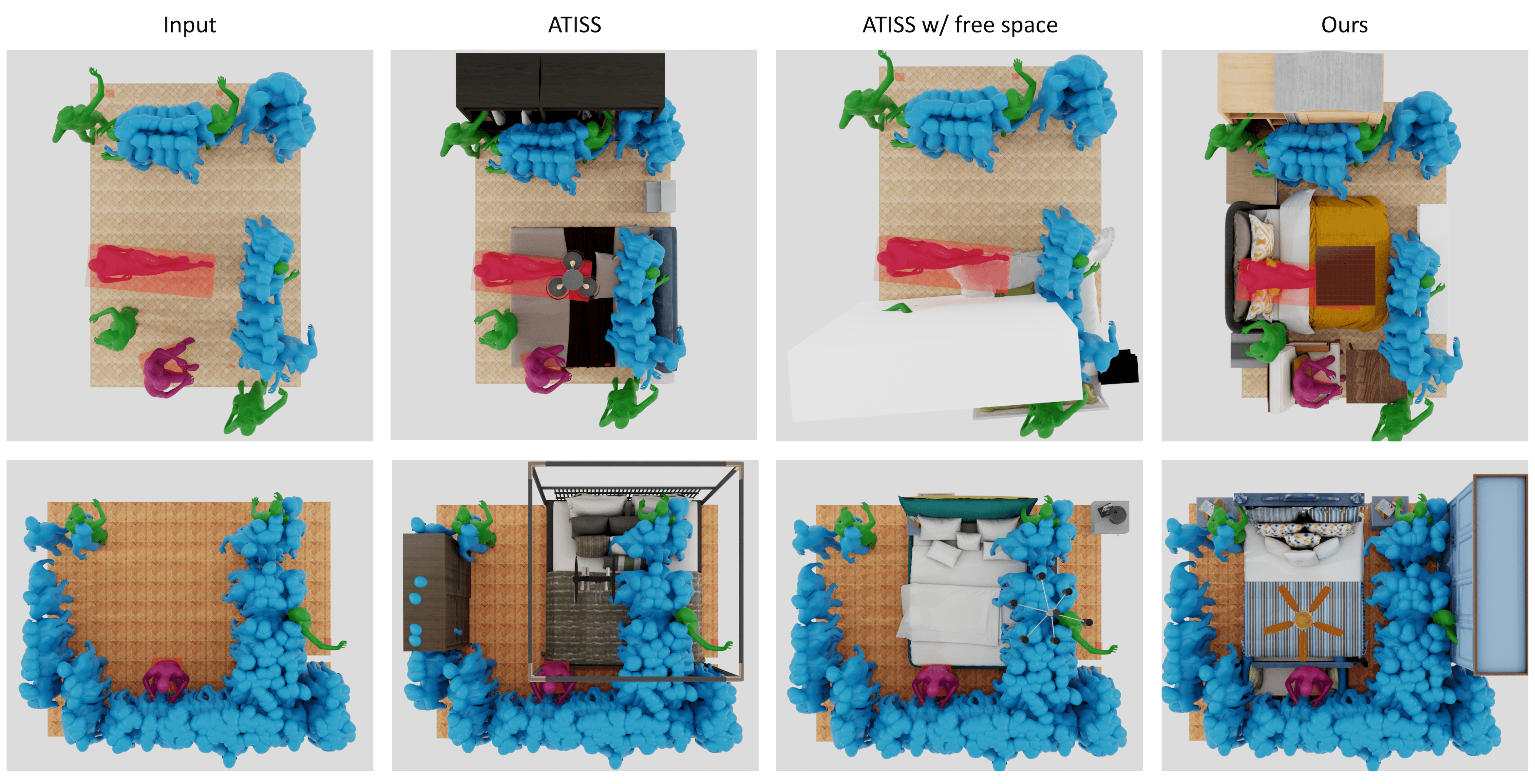}
    \vspace{0.05in}
    \caption{Qualitative comparison on the test split in \textsl{3D FRONT HUMAN}. Given free space and contact humans as input, \modelname  generates more plausible scenes in which the contact humans interact with the contact objects and the free space humans have fewer collisions with all the generated objects. We also show the original ATISS w/ or w/o the free space mask as input.
    All results are w/o refinement.
    Top and bottom rows represent two different example inputs.}
    % \hw{Blender for top-view.}} 
    \label{fig:qualitative}
\end{figure*}

%% file: tab/TAB_01_eval.tex
\begin{table*}[t]
    \centering
    \resizebox{\textwidth}{!}{
    \begin{tabular}{l|cc|cc|cc|cc|cc}
        \toprule
        \multicolumn{1}{c}{\,} & \multicolumn{2}{c}{Interpenetration($\downarrow$)} & \multicolumn{2}{c}{2D IoU($\uparrow$)}  & \multicolumn{2}{c}{3D IoU($\uparrow$)} & \multicolumn{2}{c}{FID Score ($\downarrow$)}& \multicolumn{2}{c}{Category KL Div. ($\downarrow$)}\\
        \toprule
        \multicolumn{1}{c}{\,} & ATISS \cite{Paschalidou2021NEURIPS} & Ours & ATISS \cite{Paschalidou2021NEURIPS} & Ours & ATISS \cite{Paschalidou2021NEURIPS} & Ours& ATISS \cite{Paschalidou2021NEURIPS} & Ours & ATISS \cite{Paschalidou2021NEURIPS} & Ours \\
        \midrule
        Bedroom & 0.348  &  \textbf{0.129}  &   0.472 & \textbf{0.939} & 0.376
         & \textbf{0.756}  & \textbf{70.21}$\pm$1.80  &  74.18$\pm$2.19  &  \textbf{0.028}   &    0.044 \\
        Living   &  0.129 & \textbf{0.050}  &  0.480  &  \textbf{0.971}  & 0.360 
         & \textbf{0.920}  &  \textbf{130.61}$\pm$1.27  &  150.03$\pm$ 1.00 &  \textbf{0.004}  &  0.053   \\
        Dining   & 0.121   &  \textbf{0.047}  &  0.163  & \textbf{0.959}   & 0.122
         & \textbf{0.769}  &  \textbf{45.99} $\pm$ 0.90 &  76.75 $\pm$ 1.45 &  \textbf{0.004}  &  0.037   \\
        Library  &  0.139 &  \textbf{0.106}  &  0.351  & \textbf{0.725}   & 0.390 
         & \textbf{0.570}  & \textbf{93.16} $\pm$  2.59  &    118.34$\pm$2.94  &  \textbf{0.066}  &  0.093   \\
        \bottomrule
    \end{tabular}}
    \vspace{0.05in}
        \caption{Quantitative comparison on the \textsl{test} split of the \textsl{\datasetname} dataset. The interpenetration loss, 2D IoU and 3D IoU are used to evaluate human-scene interaction in generated scenes. The FID score (reported at $256^2$) and category KL divergency are used to evaluate the realism and diversity of generated scenes, compared with ground truth scenes.}
    % Classification accuracy closer to $0.5$ is better.
    % \vspace{-1.2em}
    \label{tab:eval}
\end{table*}

%% file: figtex/FIG_06_00_comparison.tex
\begin{figure}[h]
    \centering
    \includegraphics[width=\columnwidth]{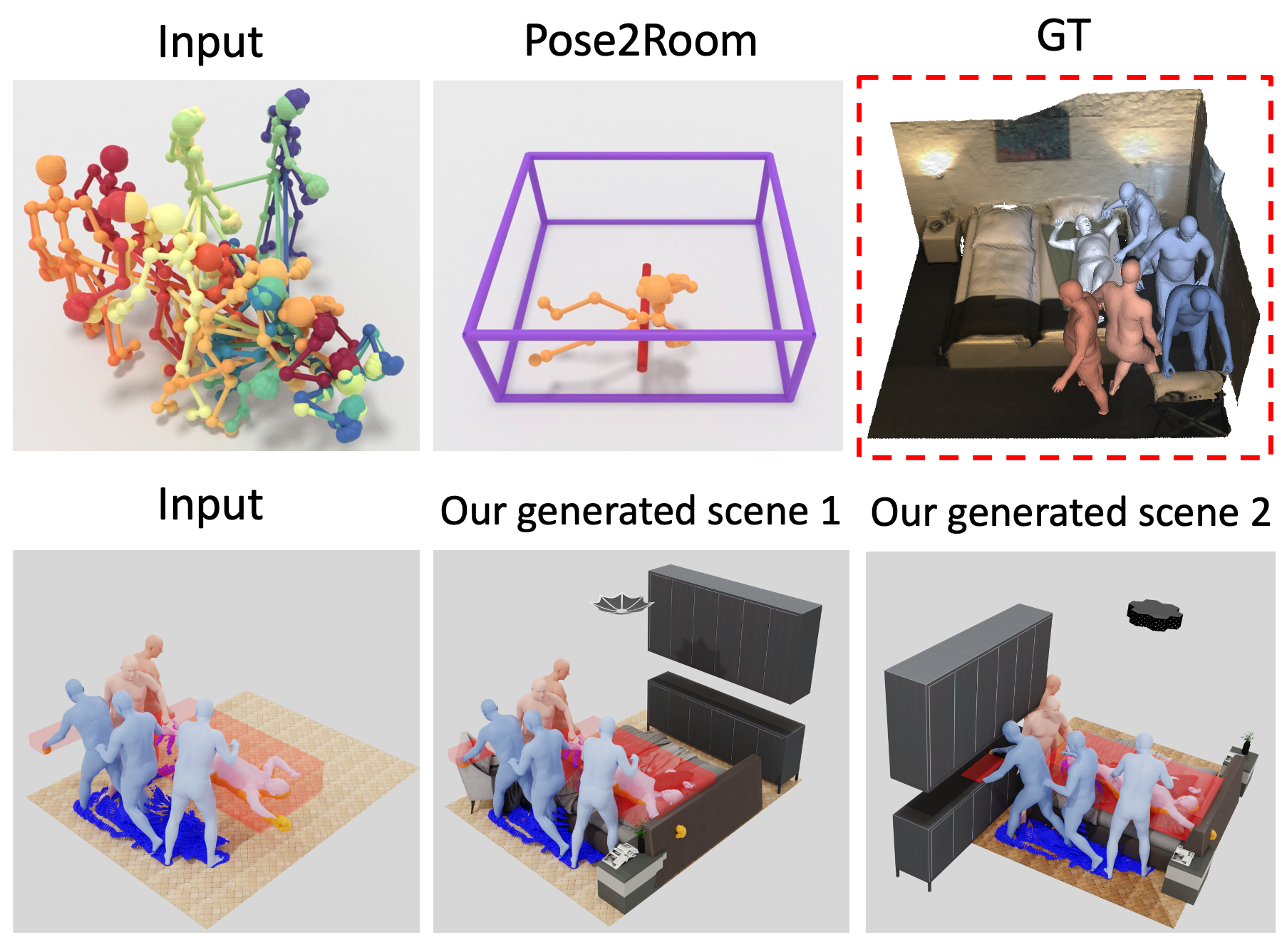}
    \vspace{0.06in}
    \caption{Evaluation on PROX \cite{hassan2019resolving, yi2022mover}. Compared with Pose2Room \cite{nie2021pose2room}, \modelname (w/o finetuning and w/o refinement) can not only generate more accurate contact objects, but it also generates objects appropriately in free space. GT = ground truth.}
    \label{fig:cmp_on_prox}
\end{figure}

%% file: tab/TAB_02_PROX.tex
% \begin{table}
% \begin{tabular}{lc||c|c|c} 
% \centering
% Method & Condition & Mean Error Frequency $\downarrow$ & More $\uparrow$ Realistic & Realism CI 99\% \\
% \hline 
% FastSynth [61] & vs. Ours & $0.414$ & $0.269$ & {$[0.235,0.306]$} \\
% SceneFormer [74] vs. Ours & $0.713$ & $0.165$ & {$[0.138,0.196]$} \\
% Ours & vs. Both & $\mathbf{0 . 2 3 2}$ & $\mathbf{0 . 7 8 3}$ & {$[0.759,0.805]$}

% \end{tabular}
% \label{tab:perceptual}
% \end{table}

\begin{table}[b!]
    \centering
    \medskip
    \resizebox{0.7\columnwidth}{!}{
    \begin{tabular}{l|c  }
            % \hline
            Method  & 3D IoU  \\ \hline
            P2R-Net \cite{nie2021pose2room} w/o pretrain & 5.36 
            \\
            Ours (\modelname) w/o pretrain & \textbf{8.47} \\ 
            % \hline
            % P2R-Net \cite{nie2021pose2room} w/ pretrain & \textbf{31.38} \\
            
            % Ours & \textbf{8.47} $\pm$ 3.68 \\ 
            
            % Ours w/ refinement & 29.30 \\
    \end{tabular}}
    \caption{Comparisons on 3D object detection accuracy (mAP@0.5) using the PROXD \textsl{qualitative} dataset \cite{hassan2019resolving}.}
    \label{tab:cmp_on_prox}
    
\end{table}

% \begin{table}[h!]
%     \centering
%     \resizebox{0.5\textwidth}{!}{
%     \begin{tabular}{@{}l@{}l||c|c|c@{}}
%     Method & Condition
%     &%
%     \begin{tabular}{@{}c@{}}
%     Mean Error \\
%     Frequency $\downarrow$ \\
%     \end{tabular}
%     &%
%     \begin{tabular}{@{}c@{}}
%     More $\uparrow$ \\
%     Realistic
%     \end{tabular}&
%     \begin{tabular}{@{}c@{}}
%     Realism \\
%     CI $99\%$
%     \end{tabular}\\
%     \hline
%     FastSynth \cite{ritchie2019fast} & \;vs. Ours & 0.414 & 0.269 & $[0.235, 0.306]$  \\
%     SceneFormer \cite{wang2020sceneformer} & \;vs. Ours & 0.713 & 0.165 & $[0.138, 0.196]$  \\
%     ATISS \cite{Paschalidou2021NEURIPS} & \;vs. Ours & 0.713 & 0.165 & $[0.138, 0.196]$  \\
%     Ours & \;vs. Both & \textbf{0.232} & \textbf{0.783} & $[0.759, 0.805]$ \\
%     \end{tabular}}
%     \captionof{table}{\textbf{Perceptual Study Results}. Aggregated results
%     for two A/B paired tests. Our method was judged more realistic
%     with high confidence (binomial confidence interval with $\alpha=0.01$ reported) and contained fewer errors.}
%     \label{tab:perceptual}
%     % \vspace{-0.8em}
% \end{table}

%% file: figtex/FIG_06_01_diversity.tex
\begin{figure}[tbh]
    \centering
    %\vspace{-0.5cm}
    \includegraphics[width=\linewidth]{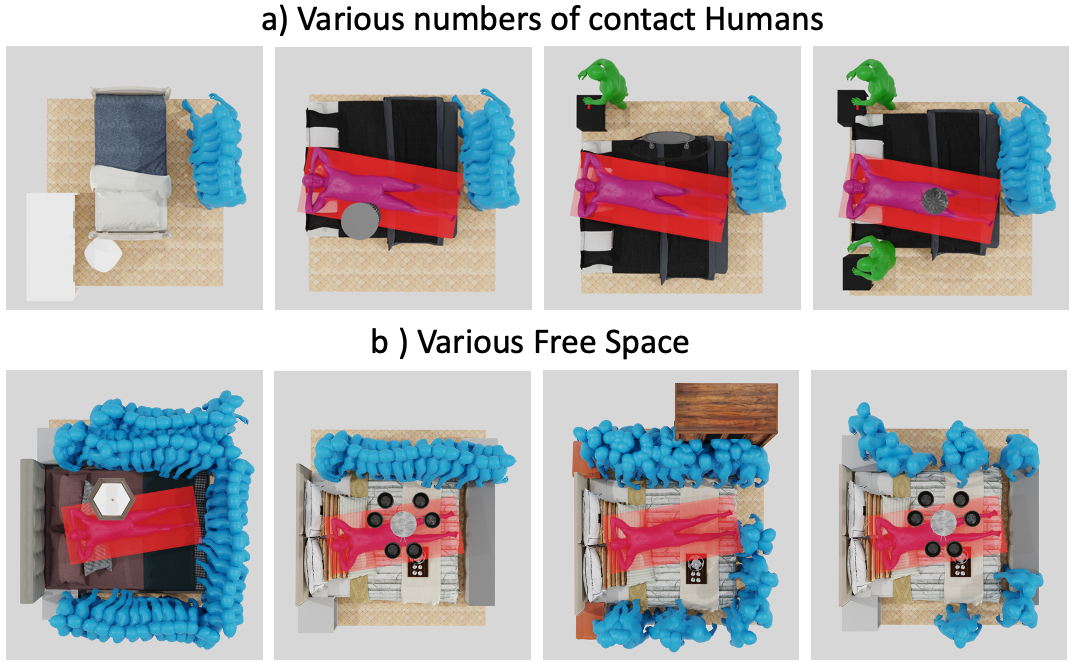}
    \vspace{0.06in}
    \caption{Ablation study on different number of contact humans and different density of free space humans.
    % \hw{add free space + one humans, evaluate the generated scene's diversity with }
    In a), with more contact humans as input, the generated scenes contain more occupied objects.
    In b), the more free space humans have in a room leads to fewer available generated objects in a scene.
    % \hw{add GT scenes.}
    % Different generated scenes conditioned on the same partial input contact humans with different free space. The different number of contact humans.
    }
    \label{fig:diversity}
\end{figure}

%% file: paper/SEC_05_discussion.tex
\section{Discussion}
Given a sequence of human motions, \modelname generates diverse and plausible scenes with which the humans interact.
We assume that the generated scenes are static, and future work should explore generating moving objects by exploring the interaction between humans and moving objects, such as moving a chair, grasping a cup, opening a door, etc.

\modelname, like ATISS, needs a pre-defined floor plan room layout as input. %with humans. 
The resolution of the 2D floor plan is coarse; i.e., 1 pixel stands for around 10 centimeters, which is extracted as a $512$ dimension feature by ResNet-18.
Introducing a finer floor plan representation, such as dividing one floor plan into multiple patches (cf.~ViT\cite{ranftl2021vision}) or simply enlarging the size of the feature dimension could improve the generated object placement, resulting in less collision between the humans and the free space. 
Another interesting direction is to estimate a floor plan and 3D object layout jointly from input humans only.

During inference, \modelname uses a hand-crafted metric 2D IoU between the generated objects and the input contact humans to factor out which human it is in contacted with. A simple extension would be to use the network to learn this information.
Our model directly estimates 3D \bboxes as a 3D scene representation, followed by a scene refinement that places the mesh models into the scene. 
Learning to directly estimate the mesh models from the interacting humans is another promising direction.

% 0. Floor plane resolution.

% 1. We need to input a pre-defined floor plane room layout. The floor plane prediction from human motion only can be one of our future work.

% 2. We make a strong assumption where one human has interaction with one object, while in real world, one human can have interactions with multiple objects, and one object can also be contacted by multiple humans.

% 3. We can not generate moving objects.
% 4. [Potential] the models are not clean, it consists of two models with the same architecture.

%% file: paper/SEC_06_conclusion.tex
\section{Conclusion}
We have introduced MIME, which generates varied furniture layouts that are consistent with input human movement and contacts.
To train MIME, we built a new dataset called \datasetname, by populating humans into the large-scale synthetic scene dataset \cite{fu20213dfront}.
We have demonstrated that by incorporating input human motion into free space and contact boxes, our method can generate multiple realistic scenes, where the input motion can take place. 
\modelname has many applications, particularly for generating synthetic training data at scale.  
\modelname provides a means of taking existing human motion capture data and ``upgrading" it to include plausible 3D scenes that are consistent with it.

%% file: paper/SEC_08_ACKOWN_DISCOL.tex
\medskip

\noindent
{\qheading{Acknowledgments.}
We thank Despoina Paschalidou, Wamiq Para for useful feedback about the reimplementation of ATISS, and Yuliang Xiu, Weiyang Liu, Yandong Wen, Yao Feng for the insightful discussions, and Benjamin Pellkofer for IT support.
This work was supported by the German Federal Ministry of Education and Research (BMBF): Tübingen AI Center, FKZ: 01IS18039B.
}

{\small \qheading{Disclosure.}
MJB has received research gift funds from Adobe, Intel, Nvidia, Meta/Facebook, and Amazon. MJB has financial interests in Amazon, Datagen Technologies, and Meshcapade GmbH. JT has received research gift funds from Microsoft Research.}

% {\qheading{Disclosure.}
% {\small
% \href{https://files.is.tue.mpg.de/black/CoI_CVPR_2022.txt}{
%       https://files.is.tue.mpg.de/black/{CoI\_CVPR\_2022.txt}}}

%% file: paper/SEC_07_appendix.tex
%%%%%%%%%%%%%%%%%%%%%%%%%%%%%%%%%%%%%%%%%%%%%%%%%%%%%%%%%%%%

% \appendix

\begin{appendices} 
\label{appendices}
\input{sup_mat/02_IMPLEM_DETAILS}

\input{sup_mat/03_MORE_EXAMPLES.tex}
\input{sup_mat/fig_tex/02_qualitative.tex}

\input{sup_mat/fig_tex/01_qualitative_01.tex}

\input{sup_mat/fig_tex/03_qualitative.tex}

\end{appendices}

%% file: sup_mat/02_IMPLEM_DETAILS.tex
\section{Training Details}

% \paragraph{Network Architecture.}

% \paragraph{Training.}
During training, we apply the Adam optimizer \cite{kingma2014adam} with learning rate $1e^{-4}$ and no weight decay.
In Adam optimizer, we use the default PyTorch implemented parameters, i.e., $\beta_{1}=0.9$, $\beta_{2}=0.999$ and $\epsilon=1e-8$.
We train MIME with the batch size $128$ for $100k$ iterations. 
We perform random global rotation augmentation between $[0, 360]$ degrees on the holistic populated scene, including the floor plane, all objects, the free space and all contact humans.
% We perform two different levels of data augmentations. One is the scene layout level, we add random rotation augmentation between $[0, 360]$ degree on the global scene layout, including the floor plane, all objects, and all humans.
% Another is the input humans level. We add translation jittering data augmentation on the free space and contact humans. 
% Given an example one 

% We train and validate MIME on the training and validation sets respectively, and evaluate it on
% the test set. 

% \paragraph{Baseline.}
% In Fig.~ of main paper, we compare our method with two baselines. One is original ATISS \cite{Paschalidou2021NEURIPS}, we 

% \paragraph{Scene refinement.}

%% file: sup_mat/03_MORE_EXAMPLES.tex
\section{More Qualitative Examples}

% In \fref{fig:qualitative}, 
We present more qualitative examples for different kinds of rooms, in \fref{fig:qualitative_bedroom}, \fref{fig:qualitative_library}, and \fref{fig:qualitative_dining}.
Compared with our baseline methods \cite{Paschalidou2021NEURIPS}, our method can generate more plausible 3D scenes that input motions can interact with. 

% the ability of our method to generate plausible 3D scenes from input motion and floor plans for different kinds of rooms; we also show our baseline methods for comparison.

%% file: sup_mat/fig_tex/02_qualitative.tex
\begin{figure}[h!]
    \centering
    \includegraphics[width=\linewidth]{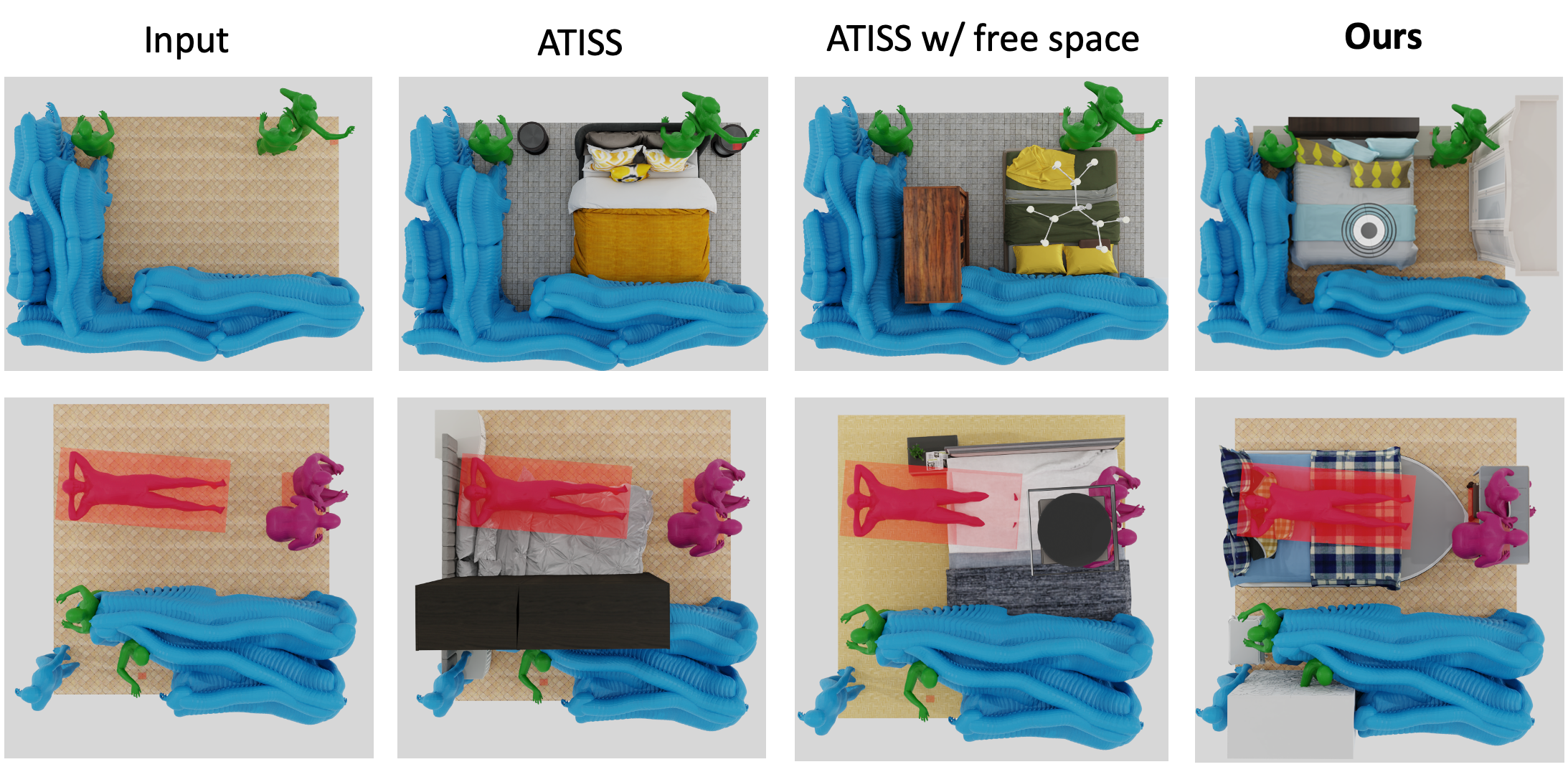}
    % \vspace{0.05in}
    
    \caption{Qualitative comparison on bedrooms in the test split of \textsl{3D FRONT HUMAN}. %
    Given free space and contact humans as input, \modelname  generates more plausible scenes in which the contact humans interact with the contact objects and the free space humans have fewer collisions with all the generated objects. We also show the original ATISS w/ or w/o the free space mask as input.
    All results are w/o refinement.
    Each row represents an example input.}
    % \hw{Blender for top-view.}} 
    \label{fig:qualitative_bedroom}
\end{figure}

%% file: sup_mat/fig_tex/01_qualitative_01.tex
\begin{figure}[htbp]
    \centering
    \includegraphics[width=\linewidth]{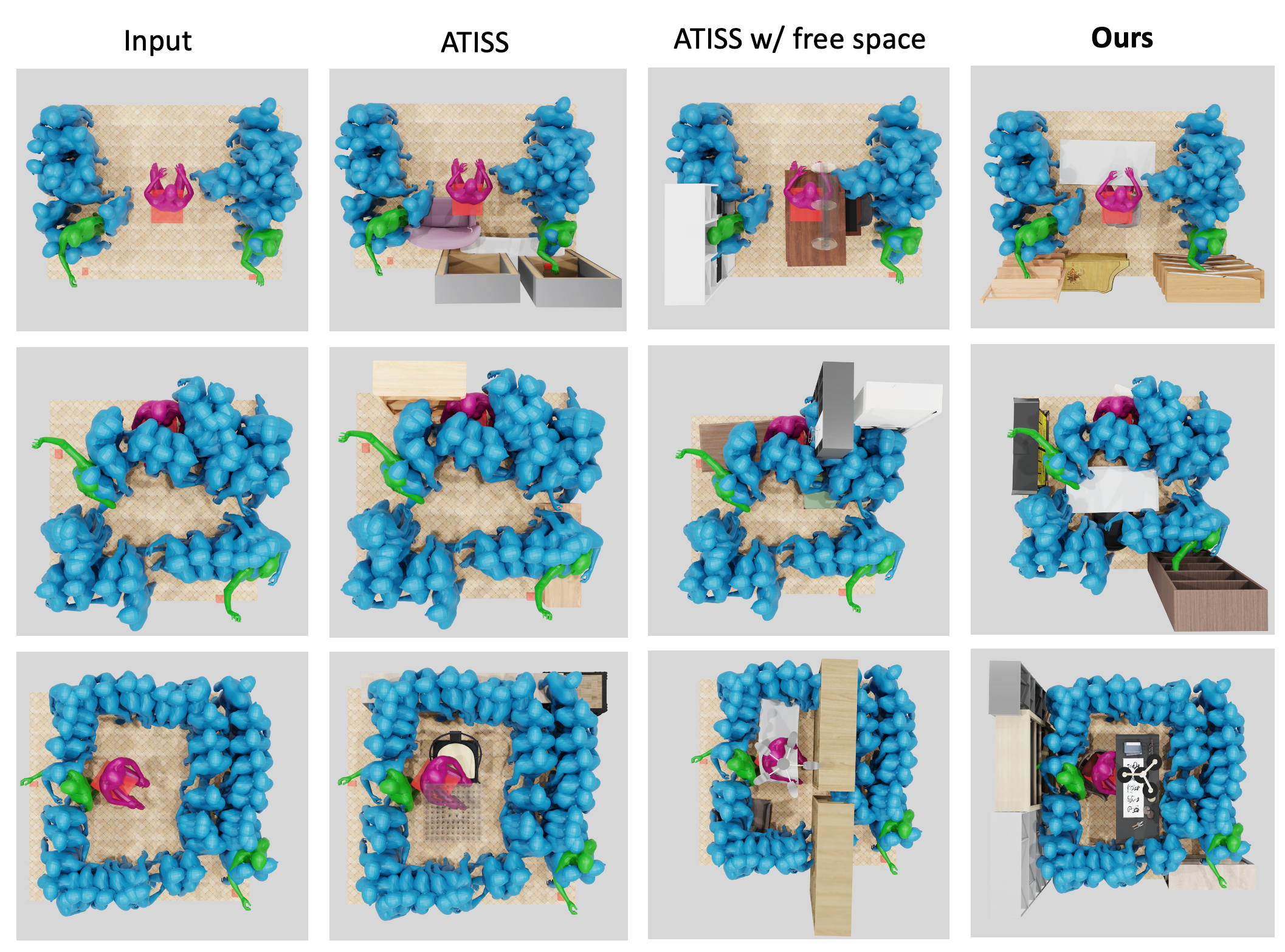}
    % \vspace{0.05in}
    \caption{Qualitative comparison on libraries in the test split of \textsl{3D FRONT HUMAN}. 
    Given free space and contact humans as input, \modelname  generates more plausible scenes in which the contact humans interact with the contact objects and the free space humans have fewer collisions with all the generated objects. We also show the original ATISS w/ or w/o the free space mask as input.
    All results are w/o refinement.
    Each row represent an example input.}
    % \hw{Blender for top-view.}} 
    \label{fig:qualitative_library}
\end{figure}

%% file: sup_mat/fig_tex/03_qualitative.tex
\begin{figure}[htbp]
    \centering
    \includegraphics[width=\linewidth]{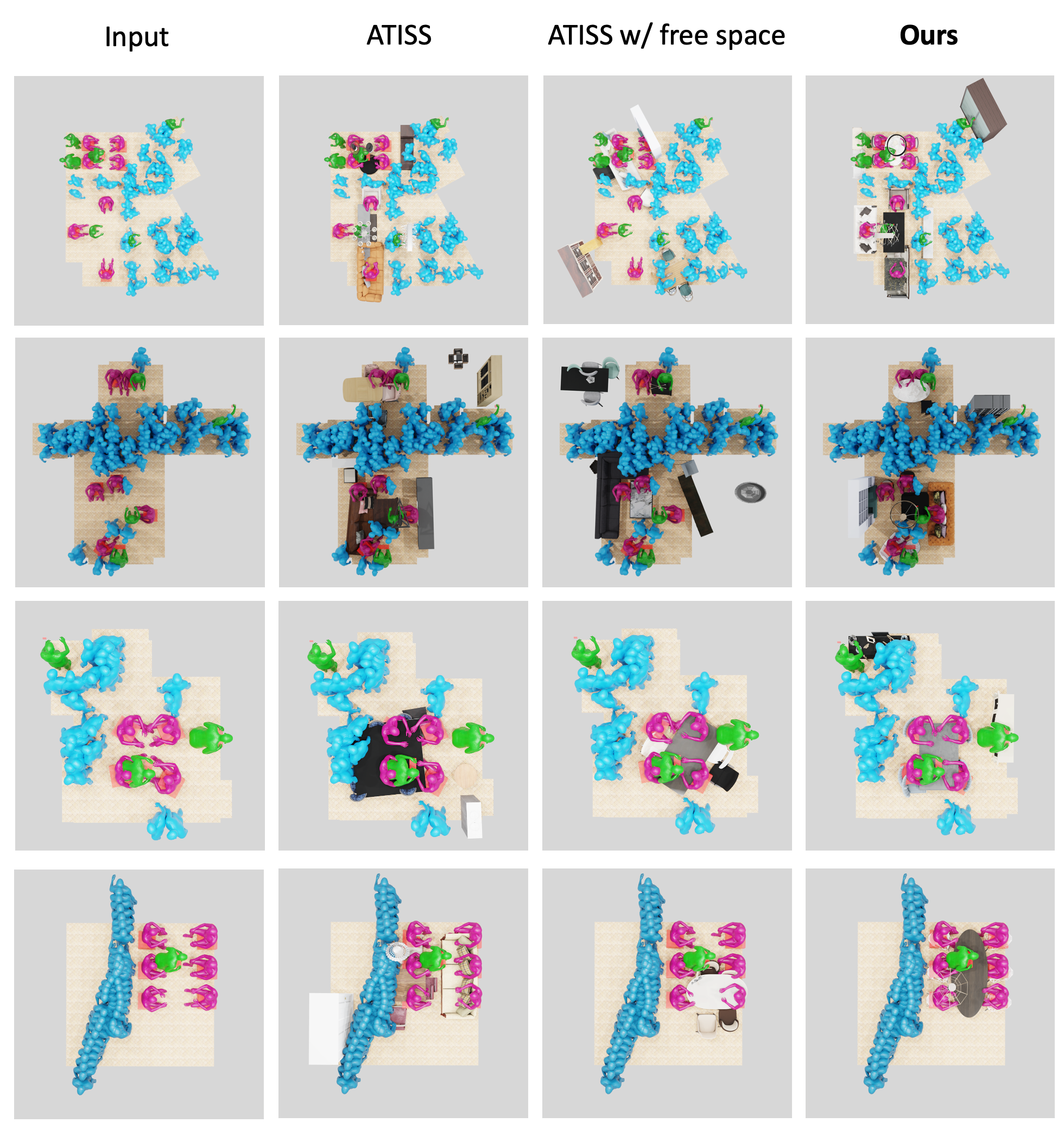}
    % \vspace{0.05in}
    \caption{Qualitative comparison on living rooms (the first two rows) and dining rooms (the last two rows) in the test split of \textsl{3D FRONT HUMAN}. 
    Given free space and contact humans as input, \modelname  generates more plausible scenes in which the contact humans interact with the contact objects and the free space humans have fewer collisions with all the generated objects. We also show the original ATISS w/ or w/o the free space mask as input.
    All results are w/o refinement.
    Each row represent an example input.}
    % \hw{Blender for top-view.}} 
    \label{fig:qualitative_dining}
\end{figure}